%% file: main.tex
\documentclass[10pt]{article}
\usepackage{tmlr}
\usepackage{graphicx}
\usepackage{cite}
\usepackage{hyperref}
\usepackage{url}
\usepackage{amsmath,amssymb,graphicx,hyperref,booktabs}
\usepackage{algorithm}
\usepackage{subcaption}
\usepackage{algpseudocode}
\usepackage{tikz}
\usepackage{multirow}
\usepackage{booktabs}
\usepackage{array}
\usepackage{placeins}
\usepackage{pgfplots}
\usepackage{natbib}
\pgfplotsset{compat=1.18}
\usetikzlibrary{shapes.geometric, arrows.meta, positioning, fit, backgrounds, calc}

\title{%
Stock Pattern Assistant (SPA): A Deterministic and Explainable Framework for Structural Price Run Extraction and Event Correlation in Equity Markets
}

\begin{document}
\maketitle

\begin{abstract}
\input{sections/abstract}
\end{abstract}

\section{Introduction}
\input{sections/introduction}

\section{Related Work}
\input{sections/relatedWork}

\section{System Overview}
\input{sections/systemOverview}

\section{Methodology}
\input{sections/methodology}

\section{Experimental Setup}
\label{sec:experimentalsetup}
\input{sections/experimentalSetup}

\section{Results: Single-Asset Structural Analysis}
\input{sections/results.tex}

\section{Event Alignment Results}
\input{sections/eventAlignmentResults}

\section{Distributional Properties and Summary Statistics}
\input{sections/distributionProperties}

\section{Cross-Asset Comparative Analysis}
\label{sec:cross_asset}
\input{sections/crossAsset}

\section{Ablation Study}
\input{sections/ablationStudy}

\section{Case Studies: Interpreting Structural Runs Through Historical Narratives}
\label{sec:case-studies}
\input{sections/interpretingCaseStudies}

\section{Run-Level Narrative Generation}
\input{sections/runLevelCaseStudies}

\section{Practical Implications}
\input{sections/practicalImplications}

\section{Robustness and Threats to Validity}
\input{sections/robustThread}

\section{Limitations}
\input{sections/limitations}

\section{Ethics Statement}
\input{sections/ethicalStatement}

\section{Reproducibility Statement}

\input{sections/reproducibility}

\section{Conclusion}

\input{sections/conclusion}

\section{Future Work}

\input{sections/futureWork}

\bibliographystyle{tmlr}
\bibliography{main}

\section*{Appendix}
\appendix
\input{sections/appendix}

\end{document}

%% file: sections/abstract.tex
Understanding how prices evolve over time often requires peeling back the layers of 
market noise to identify clear, structural behavior. Many of the tools commonly used 
for this purpose---technical indicators, chart heuristics, or even sophisticated 
predictive models---leave important questions unanswered. Technical indicators depend 
on platform-specific rules, and predictive systems typically offer little in terms of 
explanation. In settings that demand transparency or auditability, this poses a 
significant challenge.

We introduce the Stock Pattern Assistant (SPA), a deterministic framework designed to 
extract monotonic price runs, attach relevant public events through a symmetric 
correlation window, and generate explanations that are factual, historical, and 
guardrailed. SPA relies only on daily OHLCV data and a normalized event stream, making 
the pipeline straightforward to audit and easy to reproduce.

To illustrate SPA’s behavior in practice, we evaluate it across four equities---AAPL, 
NVDA, SCHW, and PGR---chosen to span a range of volatility regimes and sector 
characteristics. Although the evaluation period is modest, the results demonstrate 
how SPA consistently produces stable structural decompositions and contextual 
narratives. Ablation experiments further show how deterministic segmentation, event 
alignment, and constrained explanation each contribute to interpretability.

SPA is not a forecasting system, nor is it intended to produce trading signals. Its 
value lies in offering a transparent, reproducible view of historical price 
structure that can complement analyst workflows, risk reviews, and broader 
explainable-AI pipelines.

%% file: sections/introduction.tex
Financial markets generate rich, noisy time series that analysts must interpret 
to understand when assets experienced sustained trends, how long those runs lasted, 
what events coincided with major moves, and how patterns differ across sectors. 
Traditional technical indicators---moving averages, RSI, MACD, candlestick patterns 
\citep{murphy1999technical}---suffer from parameter sensitivity and implementation 
variance. Two charting platforms applying the same indicator often disagree on where 
trends begin and end.

Machine-learning models \citep{campbell1997econometrics, tsay2010analysis} excel at 
prediction but lack interpretability. Regulated environments such as broker-dealers 
and risk-management teams need tools that explain historical patterns, not just 
forecast future moves.

This paper introduces the \emph{Stock Pattern Assistant} (SPA), a deterministic 
framework for historical pattern extraction and explanation. SPA decomposes price 
series into monotonic runs using mathematically defined rules, aligns these runs with 
public events through a symmetric temporal window, and generates descriptive narratives 
via a constrained LLM.

We evaluate SPA on four equities spanning different sectors and volatility regimes: 
AAPL, NVDA, SCHW, and PGR. The framework produces stable decompositions without 
hyperparameter tuning for historical market structure analysis rather than trading signals.

\paragraph{Contributions.}
We introduce a deterministic segmentation framework for equity time series that requires 
no hyperparameter tuning. A symmetric event-alignment mechanism attaches contextual news 
to structural price runs. The explanation layer produces historical narratives through 
a constrained LLM. Cross-asset experiments and ablations on four equities demonstrate 
stable, interpretable decompositions, with detailed case studies supporting reproducibility.

%% file: sections/relatedWork.tex
\paragraph{Time-Series Segmentation.}
Classical approaches to segmentation include online algorithms and change-point detection 
\citep{keogh2001segmenting, truong2020changepoint}. Structural break models 
\citep{bai2003multiple, lavielle2005using} focus on regime shifts in the 
statistical properties of a series (e.g., variance, mean), rather than monotonic directional 
structure. SPA differs by targeting directional price runs that are easy to interpret and 
consistent across implementations.

\paragraph{Financial Econometrics and Market Structure.}
Foundational work in financial econometrics \citep{campbell1997econometrics, tsay2010analysis} 
and microstructure \citep{andersen2001intraday, brogaard2018price} shows that financial 
time series exhibit volatility clustering, bursts of directional movement, and regime-specific 
behavior. SPA can be viewed as a deterministic lens that exposes these bursts as monotonic 
runs, without fitting probabilistic models.

\paragraph{Event-Driven Analysis.}
Event studies \citep{mackinlay1997event, fama1998market, kothari2007event} measure 
abnormal returns around event dates, often seeking causal inference. SPA uses a 
symmetric temporal window to align runs with events for contextual correlation 
rather than causal attribution.

\paragraph{Explainability.}
Explainable AI in finance \citep{shen2022explainable, arrieta2020xai} emphasizes 
transparency. SPA's explanation layer produces descriptive, historical narratives 
grounded in deterministic run summaries and event lists \citep{lakkaraju2021interpretable}. 
Recent financial narrative generation work, including BloombergGPT, FinGPT, and LLM-based 
earnings-call summarization, produces forward-looking or sentiment-driven narratives from 
textual corpora. SPA generates backward-looking explanations grounded strictly in price 
and event structures.

\paragraph{Technical Analysis Heuristics.}
Technical analysis textbooks \citep{murphy1999technical} catalog a wide range of indicators 
and price patterns, but most lack formal, mathematically precise definitions suitable for 
reproducible analysis. SPA complements such approaches by providing a deterministic, 
parameter-light structural decomposition that can be audited and replicated.

\paragraph{Causal Inference.}
Causal inference frameworks such as synthetic controls and difference-in-differences 
isolate treatment effects in event studies. SPA focuses on correlation-based alignment 
for exploratory analysis rather than causal claims.

%% file: sections/systemOverview.tex
We intentionally provide an expanded overview to aid interpretability, 
as SPA is designed for transparency and auditability rather than 
minimalist modeling.

In this section we describe SPA’s architecture and information flow at a high level,
mirroring the modular design used in related systems.

\subsection{Components}

SPA consists of six main components:

\begin{enumerate}
  \item \textbf{Data Ingestion and Validation}
  \item \textbf{Preprocessing and Normalization}
  \item \textbf{Deterministic Run Detection}
  \item \textbf{Event Retrieval and Normalization}
  \item \textbf{Run--Event Correlation Layer}
  \item \textbf{LLM Explanation Engine}
\end{enumerate}

To situate the SPA architecture within the broader context of time-series and event-driven
analysis, it is helpful to view the system as a sequential flow of modular components, each
of which transforms raw market data into increasingly structured representations. Unlike
predictive pipelines that interleave feature engineering with model training, SPA follows a
strictly deterministic progression: raw OHLCV data are first cleaned and aligned, then
converted into signed daily returns, which in turn produce monotonic run segments with
well-defined boundaries. Public events are processed in parallel through a normalization
stage that ensures consistent timestamps, deduplication, and canonical labeling. These two
streams—structural price runs and normalized event records—are then joined by a temporal
alignment mechanism that attributes contextual information to each run without implying
causal relationships. The final explanation layer converts these structured summaries into
human-readable narratives under safety constraints. This modular design emphasizes
traceability: each representation used downstream can be mapped precisely back to inputs
from earlier stages, ensuring reproducibility and interpretability across the entire
workflow.

From a design perspective, SPA prioritizes determinism, modularity, and auditability. 
Determinism ensures that identical inputs always yield identical outputs, a property that 
distinguishes SPA from heuristic technical indicators whose behavior depends on platform-
specific implementations or hidden parameters. Modularity allows each stage of processing 
to be reasoned about independently, enabling analysts to inspect, replace, or extend 
components without altering the core segmentation logic. Auditability is achieved through 
explicit intermediate representations—cleaned price series, directional labels, monotonic 
run boundaries, event lists, and alignment outputs—each of which can be validated directly 
against raw data. This architecture supports use both as a research tool and as a 
foundation for compliance-sensitive applications that require transparent data provenance.

Technically, each module in SPA exposes a well-defined input–output interface. The 
preprocessing stage produces a sanitized OHLCV vector aligned to a unified trading 
calendar. The run-detection module takes this vector and generates a structured list of 
segments, each annotated with start and end indices, direction, duration, and 
percent change. The event-retrieval module independently outputs a normalized event 
table whose schema includes timestamps, event categories, and descriptive text. The 
alignment module joins these two streams using efficient interval-matching operations, 
returning a mapping from each run to its associated events. Finally, the explanation 
engine consumes these structured summaries and generates natural-language narratives. 
Because each interface is explicit and data-typed, the system supports modular extension: 
researchers can substitute new event sources, modify alignment rules, or add analytical 
metadata without disrupting downstream components.

For practitioners, the value of this pipeline view lies in its ability to separate 
signal from noise in a manner that remains both interpretable and reproducible. 
Market analysts often face complex, overlapping information streams—price movements, 
corporate announcements, macroeconomic updates—and must determine which historical 
patterns merit further investigation. SPA’s sequential architecture produces a 
progressively distilled representation of market behavior: raw prices become 
directionally coherent runs; unstructured news becomes a filtered, timestamp-aligned 
event set; and these components together form context-aware structural narratives. 
By transforming heterogeneous inputs into layered, human-interpretable artifacts, 
This workflow aligns closely with real-world analytical processes 
used in equity research, risk review, and compliance reporting.

\begin{figure}[h]
  \centering
  \includegraphics[width=\linewidth]{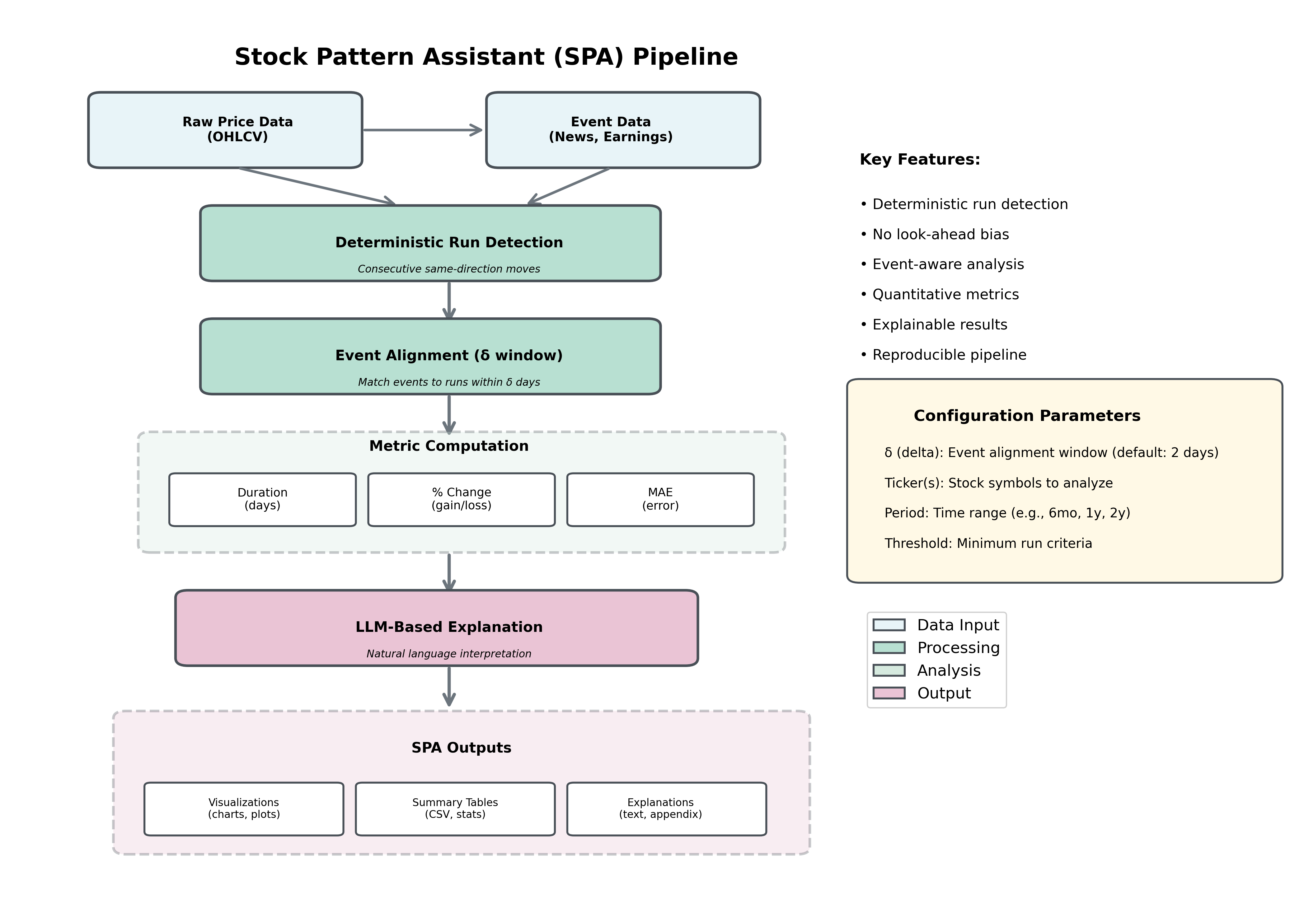}
  \caption{High-level SPA pipeline. Prices and textual events flow
    through deterministic run detection, event alignment, metric
    computation, and explanation modules to produce human-readable
    narratives for each structural regime.}
  \label{fig:spa-pipeline}
\end{figure}

Figure~\ref{fig:spa-pipeline} summarizes the end-to-end SPA pipeline,
from raw price and event ingestion through deterministic run
detection, event alignment, metric computation, and LLM-based
explanation.

\FloatBarrier

%% file: sections/methodology.tex
We decompose financial time series into monotonic directional runs, align them with 
public events, and generate narrative summaries. This section formalizes each stage.

\subsection{Data Model and Notation}
\label{sec:data-model}

Let $\{P_t\}_{t=1}^T$ denote the adjusted close prices for a given asset over $T$ trading
days. We assume prices are aligned to a business-day calendar with adjustments for
splits and dividends. SPA starts by computing daily differences:
\[
\Delta_t = P_t - P_{t-1}, \qquad t = 2,\dots,T,
\]
and assigning a directional label
\[
d_t =
\begin{cases}
+1, & \Delta_t > 0,\\[4pt]
-1, & \Delta_t < 0,\\[4pt]
0,  & \Delta_t = 0.
\end{cases}
\]

Days with $d_t=0$ represent flat moves and are treated as neutral separators that do not
contribute to directional runs. The sequence $\{d_t\}$ forms the foundation for 
deterministic segmentation. Using only the sign of price differences---rather than 
thresholds, smoothing windows, or statistical filters---ensures determinism, 
implementation invariance, and robustness under scaling or affine transformations.

\subsection{Deterministic Run Extraction}
\label{sec:run-extraction}

A \emph{monotonic run} is a maximal contiguous interval $R_k = [s_k, e_k]$ such that
\[
d_t \in \{+1, -1\} \quad \text{and} \quad d_t = d_{t+1} \quad \forall\ t\in[s_k, e_k - 1].
\]
Boundary conditions ensure maximality:
\[
d_{s_k-1} \neq d_{s_k} \quad \text{and} \quad d_{e_k+1} \neq d_{e_k},
\]
with appropriate handling at the edges of the time series.

For each run $R_k$ we compute:
\[
\text{Duration}(R_k) = e_k - s_k + 1,
\qquad
\text{PctChange}(R_k) = \frac{P_{e_k} - P_{s_k}}{P_{s_k}} \times 100\%.
\]

To quantify intra-run stability, SPA computes a mean absolute excursion measure:
\[
\text{MAE}(R_k) = 
\min_{t \in [s_k,e_k]} \frac{P_t - P_{s_k}}{P_{s_k}}.
\]
Because SPA requires monotonic price behavior, $P_t \ge P_{s_k}$ for upward runs and 
$P_t \le P_{s_k}$ for downward runs, yielding $\text{MAE}(R_k) = 0$ as a diagnostic 
property. Run extraction requires a single pass over $\{d_t\}$ in $O(T)$ time and 
$O(K)$ memory, where $K$ is the number of runs.

\subsection{Event Retrieval and Normalization}
\label{sec:event-normalization}

SPA ingests a collection of timestamped events
\[
E = \{(\tau_j, \ell_j, c_j)\}_{j=1}^M,
\]
where $\tau_j$ is the event timestamp, $\ell_j$ is an event category (e.g., “Earnings”,
“Product Launch”), and $c_j$ is a short textual description.

To ensure consistency, SPA performs:
\begin{itemize}
    \item \textbf{Timestamp alignment}: convert timestamps to a unified timezone and
          business-day calendar.
    \item \textbf{Deduplication}: merge or drop redundant events using fuzzy matching on
          text and categories.
    \item \textbf{Schema standardization}: normalize fields so downstream modules do not
          depend on vendor-specific formats.
\end{itemize}

Normalization is deterministic and reproducible, ensuring that identical event feeds
produce identical aligned outputs.

\subsection{Run--Event Temporal Alignment}
\label{sec:event-alignment}

SPA defines a symmetric correlation window around each run:
\[
W_k = [s_k - \delta,\, e_k + \delta],
\]
where $\delta$ is typically 1–3 trading days. An event $(\tau_j, \ell_j, c_j)$ is aligned
with run $R_k$ if:
\[
\tau_j \in W_k.
\]

Several design properties follow from this definition:

\paragraph{Non-causal interpretation.}
SPA does not infer that events \emph{cause} price movements; the window captures temporal
co-occurrence only. This satisfies compliance constraints for regulated environments.

\paragraph{Event-density diagnostics.}
Event alignment supports computation of:
\[
\text{Density}(R_k) = \frac{\# \text{aligned events}}{\text{Duration}(R_k)}.
\]
High-density runs often signal informationally rich market regimes.

\paragraph{Computational Efficiency.}
Using a sorted event list, event alignment can be implemented via two-pointer sweeping or
binary search, giving:
\[
O(K \log M) \quad \text{or even} \quad O(K+M)
\]
depending on choice of implementation.

A quantitative sensitivity analysis appears in Appendix~\ref{sec:appendix-D}.

\subsection{Explanation Layer and Guardrails}
\label{sec:explanation-layer}

SPA’s final stage converts each structured run summary into a natural-language narrative.
The input to the LLM is a JSON object containing:
\begin{itemize}
    \item ticker symbol,
    \item run direction and duration,
    \item start and end dates,
    \item list of aligned events $(\tau_j, \ell_j, c_j)$ with standardized categories.
\end{itemize}

To guarantee safety and reproducibility, the LLM operates under strict guardrails:

\paragraph{No forward-looking statements.}
The system prompt explicitly forbids speculation or prediction.

\paragraph{No causal attribution.}
The LLM must phrase event references in descriptive terms (“coincided with”,
“occurred during”) rather than causal ones (“caused”).

\paragraph{Consistency constraints.}
Narratives must reference only provided fields; no external knowledge may be injected.

\paragraph{Deterministic inputs.}
Because run segmentation and event alignment are deterministic, explanations are as
stable as the underlying data, and LLM stochasticity affects only phrasing, not content.

\paragraph{Auditability.}
All inputs and outputs are logged and reproducible, enabling downstream scoring,
interpretation, and compliance review.

\subsection{Methodological Advantages and Comparison to Existing Approaches}
\label{sec:method-comparison}

SPA occupies a methodological space distinct from both traditional technical-analysis
heuristics and statistically driven segmentation procedures. In this subsection we 
clarify these differences and articulate why SPA’s deterministic structure is uniquely 
well-suited for transparent financial analysis.

\paragraph{Contrasts with Technical Indicators.}
Most classical technical indicators—such as moving averages, RSI, MACD, stochastic 
oscillators, or Bollinger Bands—depend on hyperparameters, smoothing windows, or 
platform-specific implementation choices. Small changes in these parameters often 
produce large differences in trend signals, undermining reproducibility. Moreover, 
technical indicators rarely yield clean, interpretable units of structure: a crossover 
event or overbought/oversold condition does not map directly to a well-defined segment 
of the price series. By contrast, SPA produces explicit monotonic segments whose 
boundaries follow a mathematically precise and platform-invariant definition. Each run 
is an interpretable atomic unit representing a sustained directional phase of price 
movement.

\paragraph{Contrasts with Change-Point Detection.}
Classical change-point and structural break models 
~\citep{bai2003multiple,lavielle2005using,truong2020changepoint}
identify shifts in underlying statistical regimes—for example, jumps in volatility or 
changes in mean. These methods typically optimize criteria such as penalized likelihoods 
or contrast functions. While powerful, they do not inherently target monotonic structure 
and may group numerous directional shifts into a single statistical regime. SPA instead 
focuses on \emph{direction-consistent} segments, providing a complementary structural 
decomposition capturing micro-regimes of sustained accumulation or distribution.
Unlike change-point detection, SPA does not require hyperparameter tuning (e.g., 
penalty strength, minimum segment length), and it runs in strictly linear time.

\paragraph{Contrasts with Hidden Markov Models and Regime-Switching Approaches.}
Regime-switching models such as Hidden Markov Models (HMMs) 
~\citep{hamilton1989regime} infer latent states governing price evolution. While 
these models can capture transitions between high-volatility and low-volatility regimes, 
their inferred states depend heavily on initialization, model specification, and 
likelihood optimization. They provide probabilistic regime assignments rather than 
deterministic structural units. SPA avoids such ambiguity: runs are defined purely 
by observed directional structure and are identical across implementations, making 
them suitable for audit and compliance workflows where reproducibility is essential.

\paragraph{Contrasts with Predictive Machine-Learning Pipelines.}
Modern predictive models—including deep learning architectures and feature-driven 
supervised learning systems \citep{gu2020deeplearning}—offer strong forecasting ability 
but typically lack interpretability. Their internal representations (e.g., embeddings, 
attention maps) do not correspond to semantically meaningful historical structures. 
SPA inverts this paradigm by focusing entirely on deterministic pattern extraction 
without prediction. The outputs produced by SPA (runs, event alignments, and 
narratives) are directly interpretable, making them suitable as transparent diagnostics, 
feature inputs for downstream supervised models, or building blocks for multi-layer 
explainability frameworks.

\paragraph{Advantages of SPA’s Deterministic Architecture.}
The deterministic nature of SPA confers several benefits:
\begin{itemize}
    \item \textbf{Reproducibility:} identical inputs always produce identical outputs.
    \item \textbf{Auditability:} intermediate structures (e.g., run boundaries, event 
          alignments) can be traced directly to raw data.
    \item \textbf{Parameter-free design:} no smoothing parameters, penalty weights, or 
          thresholds are required.
    \item \textbf{Computational simplicity:} the run-extraction algorithm is $O(T)$ and 
          practical for large universes and intraday extensions.
    \item \textbf{Interpretability:} each run represents a semantically meaningful unit 
          of sustained directional movement.
\end{itemize}

Stepping back, SPA clearly serves a purpose that the other methods do not. Its output is not a softened version of technical 
indicators or a probabilistic regime label. It is a deliberately simple, clearly defined 
structural lens, whose value comes from determinism and interpretability.

The full system prompt is provided in Appendix~\ref{sec:appendix-E}.

%% file: sections/experimentalSetup.tex
\subsection{Assets and Time Horizon}

We evaluate SPA on four equities:
AAPL (mega-cap technology), 
NVDA (high-volatility semiconductor), 
SCHW (financial services), 
and PGR (insurance).  

For each, we retrieve approximately six months of daily OHLCV data from Yahoo Finance 
\citep{yahoofinance}. We adjust for splits and dividends when applicable and verify 
the continuity of the trading calendar.

For reproducibility and deterministic evaluation, events were generated from 
curated, timestamped sample data representing typical corporate announcements, 
analyst commentary, and newsworthy developments for each ticker. All timestamps 
are standardized to U.S. Eastern Time. The implementation supports configurable 
event sources including Alpha Vantage News Sentiment API and NewsAPI.org for 
live production use.

\subsection{Preprocessing and Implementation Details}

We implement SPA in Python using standard data science libraries. We:

\begin{itemize}
  \item align timestamps to a unified trading calendar,
  \item drop days with missing or anomalous prices,
  \item compute $\Delta_t$ and directional labels $d_t$,
  \item run the deterministic run-detection algorithm,
  \item fetch and normalize events, and
  \item generate all figures and tables via a reproducible script.
\end{itemize}

The implementation includes an experiment orchestration module that executes the 
pipeline end-to-end: data download, preprocessing, run extraction, event alignment, 
and figure generation. All outputs are written to a standardized directory structure, enabling
reviewers and practitioners to regenerate the full set of results from raw data with
a single command.

%% file: sections/results.tex
\subsection{AAPL: Balanced Structural Momentum in a Mega-Cap Environment}

AAPL displays a relatively symmetric distribution of upward and downward runs. Run lengths 
concentrate around 2--3 days, with occasional multi-day trends extending to 6--7 days. Both 
upward and downward sequences are smoother than NVDA's, with fewer micro-fragmented reversals, 
reflecting the stock's large market capitalization and broad investor base.

The percent-change distribution shows upward runs cluster around 1--3\% gains, while downward 
runs show slightly larger losses but occur less frequently. Event density is moderate, with 
aligned events typically being earnings releases, product updates, or buyback announcements. 
Appendix~\ref{sec:appendix-C} examples show cases where AAPL experienced strong gains absent significant events.

\subsection{Cross-Asset Event Analysis}

Event-alignment plots in Figure~\ref{fig:event_alignment} and summary statistics in 
Table~\ref{tab:summary} show systematic differences. NVDA exhibits shorter, sharper upward 
bursts, while AAPL and PGR show longer mixed regimes where events cluster around both run 
starts and reversals. Event patterns differ by sector:

\begin{itemize}
  \item \textbf{Technology (NVDA, AAPL):} earnings, product launches, analyst upgrades.
  \item \textbf{Financials (SCHW):} rate decisions, regulatory commentary, macro releases.
  \item \textbf{Insurance (PGR):} partnership announcements, actuarial updates, 
        product-cycle shifts.
\end{itemize}

The symmetric event window shows how certain runs consistently overlap with event clusters, 
revealing whether structural moves occur in informationally dense environments or arise from 
technical dynamics.

\subsection{NVDA: High-Volatility Fragmentation and Momentum-Driven Regimes}

NVDA's price history is characterized by short, intense, and frequently alternating 
directional regimes. The stock exhibits the largest number of runs among all four tickers, 
with heavy concentration of 1--3 day monotonic segments. This burst-like momentum reflects 
high-beta technology stocks where localized demand shocks quickly reverse due to profit-taking 
or sentiment changes.

A notable example is a 6-day upward run with 9.59\% gain (Appendix~\ref{sec:appendix-C}), representing one 
of the strongest multi-day structural moves in the dataset. NVDA's upward runs show both 
higher average magnitude and larger maximum gains than other tickers, while downward runs 
exhibit correspondingly deeper losses. This asymmetry aligns with rapid appreciation during 
positive technology sentiment and sharp declines during valuation resets.

Figure~\ref{fig:event_alignment} shows NVDA has higher event density relative to other 
assets. Aligned events often include product announcements, semiconductor supply-chain news, 
and analyst rating revisions. The event-alignment mechanism surfaces contextual patterns 
that would remain hidden in purely price-based analysis.

\subsection{SCHW: Macro-Sensitive Behavior and Policy-Driven Run Dynamics}

SCHW's price behavior is heavily influenced by interest-rate announcements, regulatory shifts, 
and macroeconomic data. The run-length distribution reveals both short sequences driven by 
daily market sentiment and longer multi-day regimes associated with macro cycles. Compared to 
AAPL, SCHW displays greater variability in run length, reflecting retail flows, institutional 
rebalancing, and rate expectations.

Downward runs in SCHW often coincide with policy uncertainty or macroeconomic stress windows. 
Upward runs typically emerge during interest-rate stability or supportive regulatory developments. 
Figure~\ref{fig:event_alignment} shows events often cluster in proximity to directional shifts. 
Policy announcements and financial-sector commentary tend to align with the onset of downward runs.

\subsection{PGR: Insurance-Sector Persistence and Extended Structural Regimes}

PGR displays some of the longest runs among all evaluated equities. Multi-day downward 
runs indicate sustained pressure that often 
corresponds to cyclical insurance factors, actuarial updates, or premium-adjustment periods. 
Unlike NVDA or AAPL, where run fragmentation dominates, PGR's segmentation reveals clear, 
persistent momentum regimes.

PGR's overall volatility remains lower relative to technology names, producing more stable 
percent-change distributions and smoother run-length histograms. Figure~\ref{fig:event_alignment} 
shows PGR's aligned events are fewer but more sector-specific. Notable alignments include 
partnership announcements and product launches that often coincide with sharp structural moves.

%% file: sections/eventAlignmentResults.tex
\begin{figure}[ht]
\centering
% four-panel event alignment
\begin{subfigure}{0.48\textwidth}
\includegraphics[width=\linewidth]{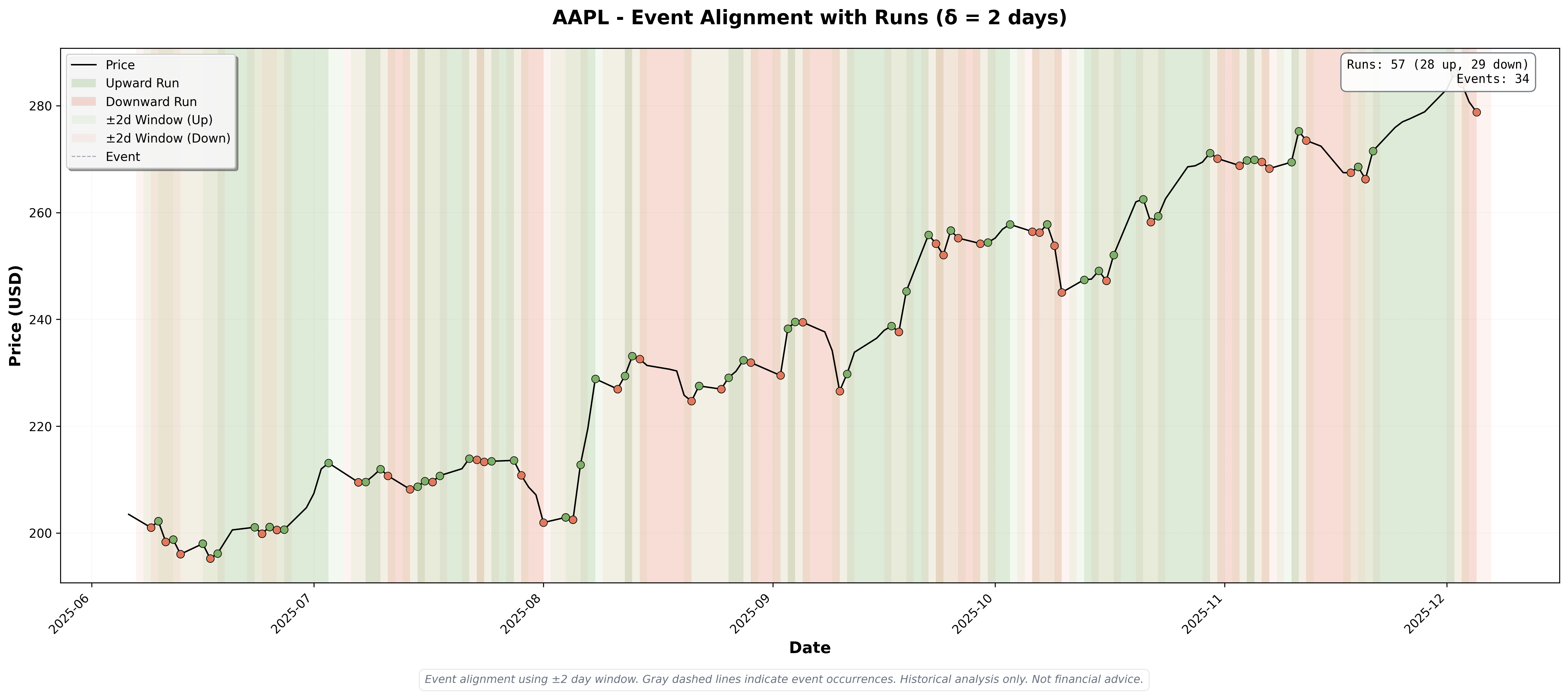}
\caption{AAPL}
\end{subfigure}
\begin{subfigure}{0.48\textwidth}
\includegraphics[width=\linewidth]{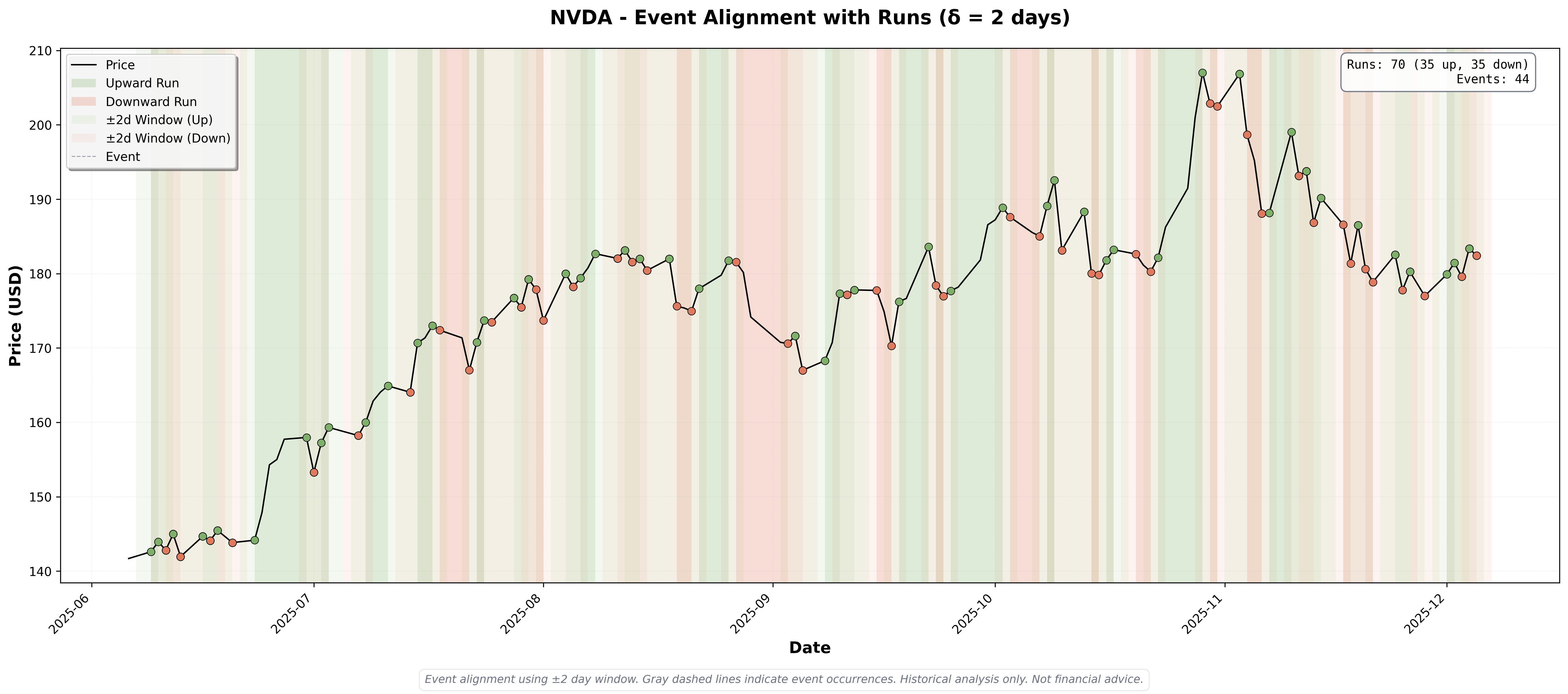}
\caption{NVDA}
\end{subfigure}

\begin{subfigure}{0.48\textwidth}
\includegraphics[width=\linewidth]{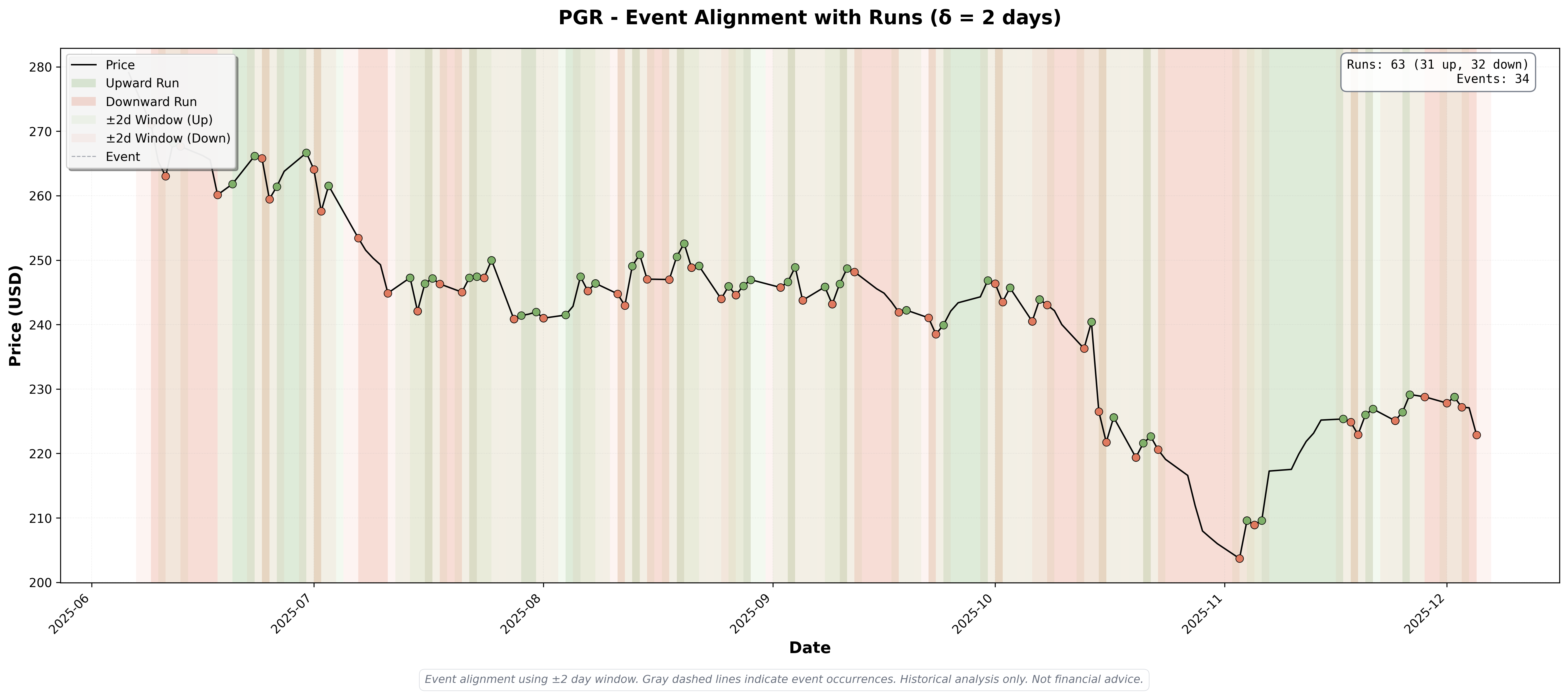}
\caption{PGR}
\end{subfigure}
\begin{subfigure}{0.48\textwidth}
\includegraphics[width=\linewidth]{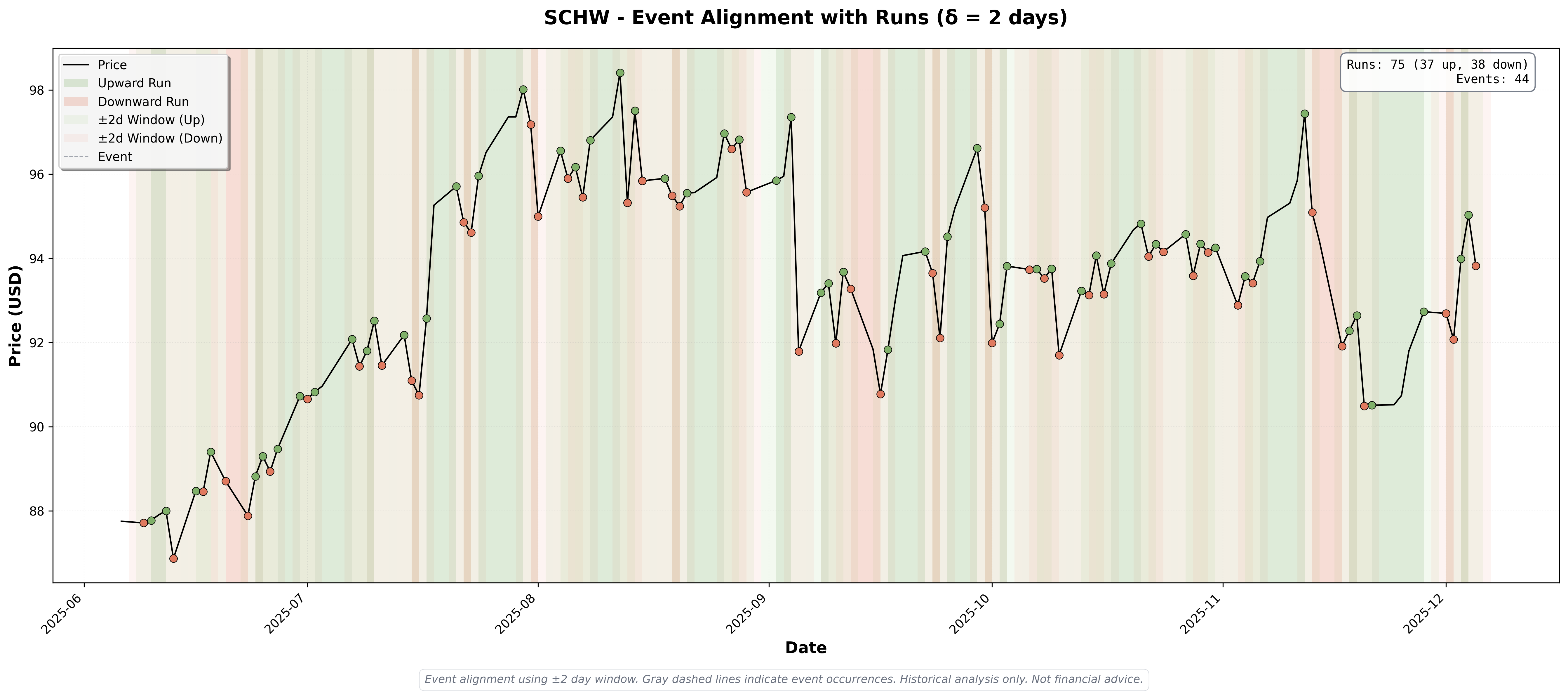}
\caption{SCHW}
\end{subfigure}
\caption{
Event alignment with directional runs using $\delta=2$. Gray markers denote public events temporally correlated with each run.
}
\label{fig:event_alignment}
\end{figure}
\FloatBarrier

Event alignment shows meaningful patterns in how public information clusters around structural
price runs. Across the four tickers, Figure~\ref{fig:event_alignment} shows that the temporal
overlap between runs and events varies systematically across sectors, providing insight into how
information flow interacts with directional market structure.

For AAPL, events tend to be sparse but appear near the beginning or end of multi-day runs, 
highlighting the influence of product updates and earnings cycles. These event clusters often
coincide with transitions between upward and downward segments, suggesting that public disclosures
shape sentiment adjustments even when they do not directly cause price movements.

NVDA exhibits the highest event density among the four assets. The semiconductor and AI sectors
generate frequent analyst commentary, supply-chain news, and product announcements. This is 
reflected in dense clusters of gray event markers throughout the run sequence. Notably, NVDA shows
multiple runs that occur in close proximity to several events, although SPA’s conservative design
avoids implying any causal relationship. Instead, the mechanism produces a structured temporal correlation that
helps contextualize NVDA’s volatility.

PGR, an insurance-sector equity, shows fewer events but stronger alignment when events do occur. 
Sector-specific news such as partnership announcements or actuarial updates often coincide with long
downward runs. This pattern reflects slower-moving fundamentals in the insurance industry and 
shows how event alignment can expose latent drivers of structural persistence.

SCHW shows alignment patterns that are clearly macro-sensitive. Policy announcements, commentary
related to interest rates, and financial-sector regulatory updates frequently coincide with directional
changes. For instance, downward runs often overlap with windows surrounding macro announcements,
reflecting the firm’s sensitivity to changes in rate expectations.

Together, These patterns reinforce a point worth emphasizing: SPA is most useful as an 
interpretive layer. It helps the analyst notice when a run unfolded in a period full 
of public information versus when it emerged quietly, without any clear event 
pressure. It does not claim to forecast or even infer why these moves happened. The $\delta$-window mechanism indicates whether a run occurred in an information-rich 
interval or an information-sparse phase, equipping analysts and auditors with a compliance-safe and 
reproducible interpretation of historical market behavior.

%% file: sections/distributionProperties.tex
\begin{figure}[h]
\centering
\includegraphics[width=\linewidth]{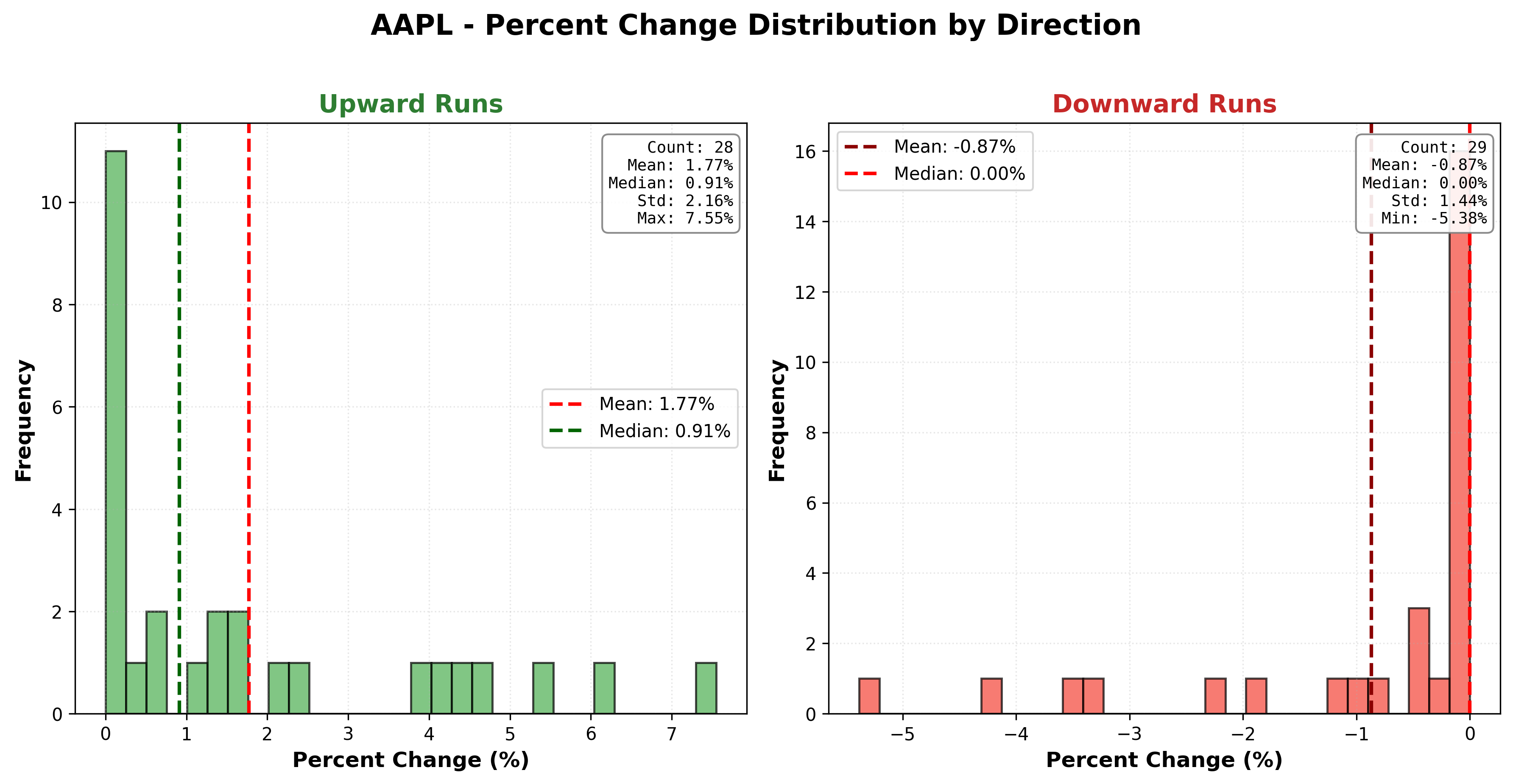}
\caption{Percent-change distributions for AAPL upward and downward runs.}
\label{fig:aapl_pct_hist}
\end{figure}
\FloatBarrier

\begin{figure}[h]
\centering
\includegraphics[width=\linewidth]{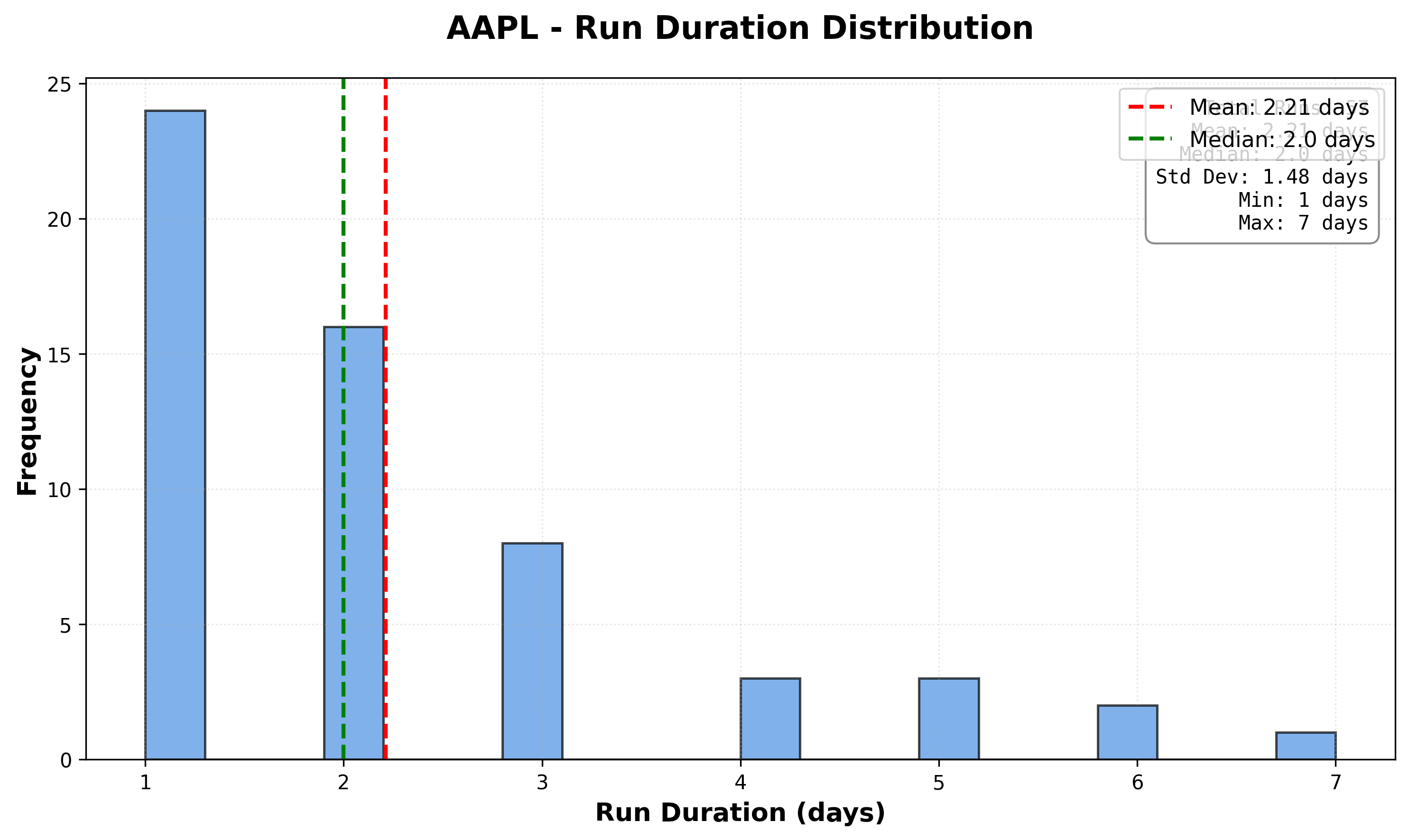}
\caption{Distribution of AAPL run durations.}
\label{fig:aapl_durations}
\end{figure}
\FloatBarrier

\begin{table}[h]
  \centering
  \caption{Run statistics across tickers over the June--December 2025
  sample. Durations are in trading days. Max gain/loss are the largest
  peak-to-trough percentage changes observed within a single run.}
  \label{tab:summary}
  \begin{tabular}{lrrrrrrrr}
    \toprule
Ticker & \# Runs & \# Up & \# Down & Mean dur & Median dur & Max gain & Max loss \\
       &         &      &         & (days)   & (days)      & (\%)     & (\%)      \\
    \midrule
    AAPL & 57 & 28 & 29 & 2.21 & 2.00 & 7.55 & -5.38 \\
    NVDA & 70 & 35 & 35 & 1.80 & 1.00 & 13.66 & -6.05 \\
    SCHW & 75 & 37 & 38 & 1.67 & 1.00 & 3.73 & -3.37 \\
    PGR  & 63 & 31 & 32 & 2.00 & 2.00 & 7.51 & -7.66 \\
    \bottomrule
  \end{tabular}
\end{table}
\FloatBarrier

\begin{figure}[h]
\centering
% four-panel run detection figure
\begin{subfigure}{0.48\textwidth}
\includegraphics[width=\linewidth]{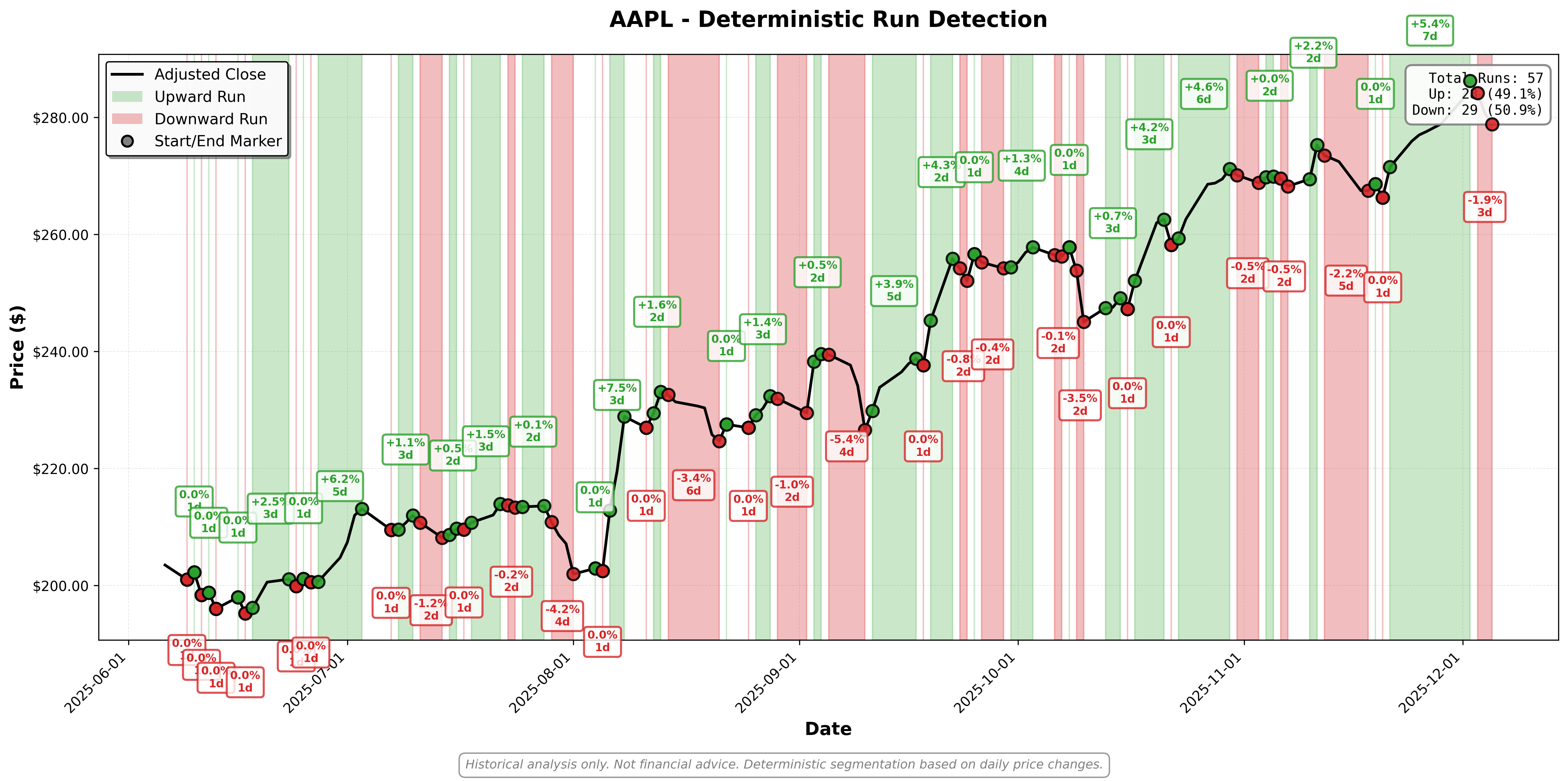}
\caption{AAPL}
\end{subfigure}
\begin{subfigure}{0.48\textwidth}
\includegraphics[width=\linewidth]{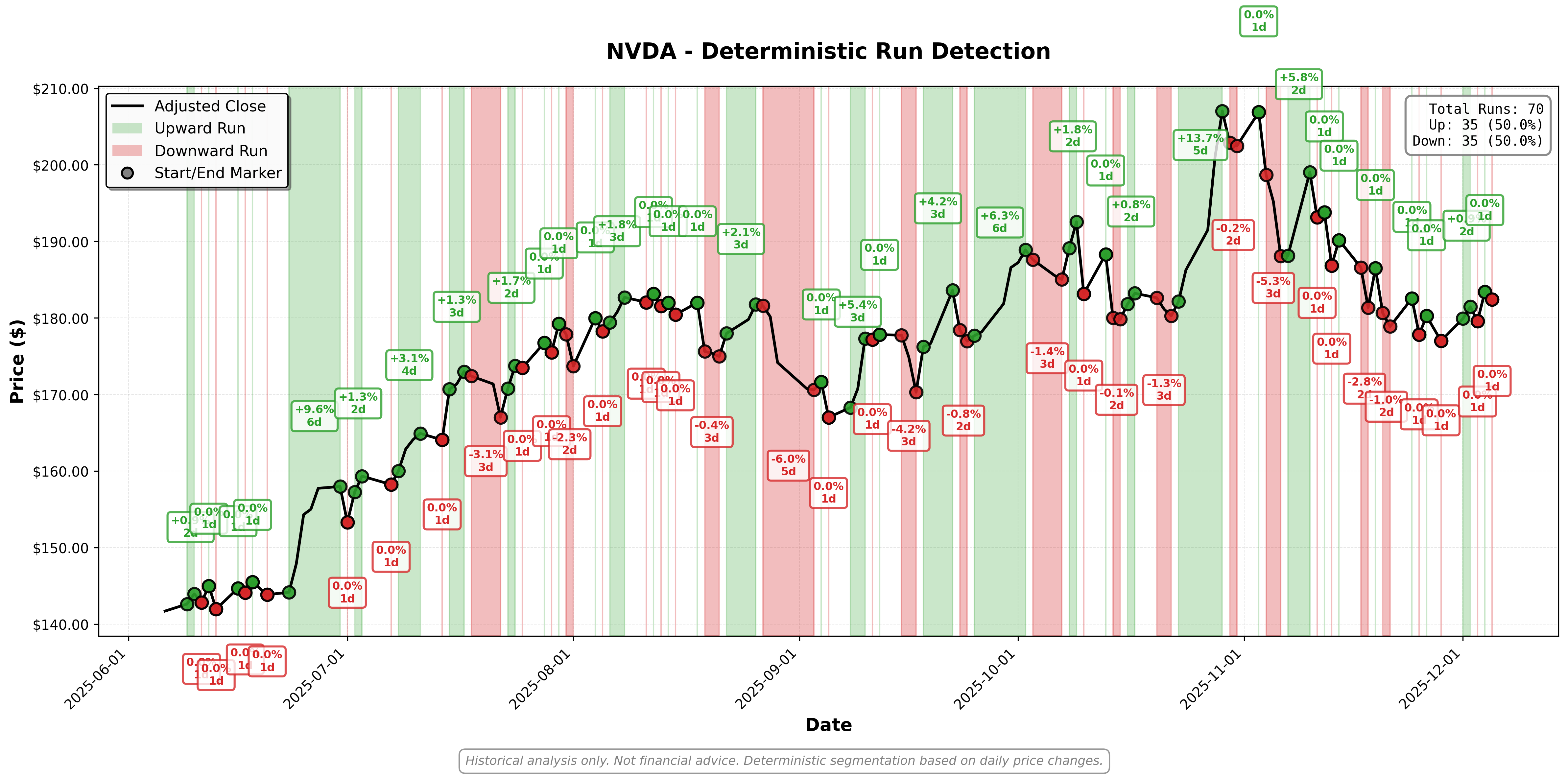}
\caption{NVDA}
\end{subfigure}

\begin{subfigure}{0.48\textwidth}
\includegraphics[width=\linewidth]{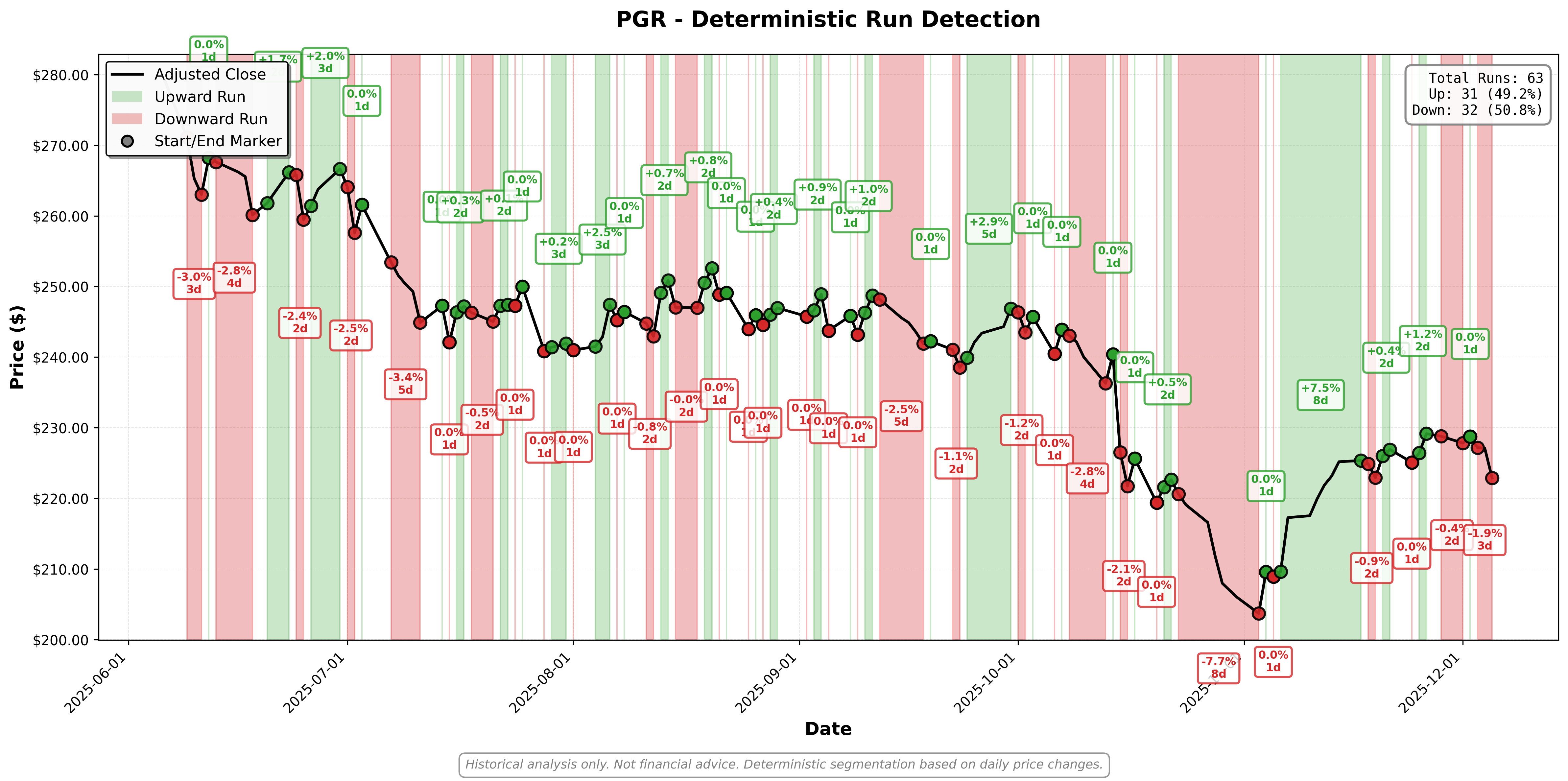}
\caption{PGR}
\end{subfigure}
\begin{subfigure}{0.48\textwidth}
\includegraphics[width=\linewidth]{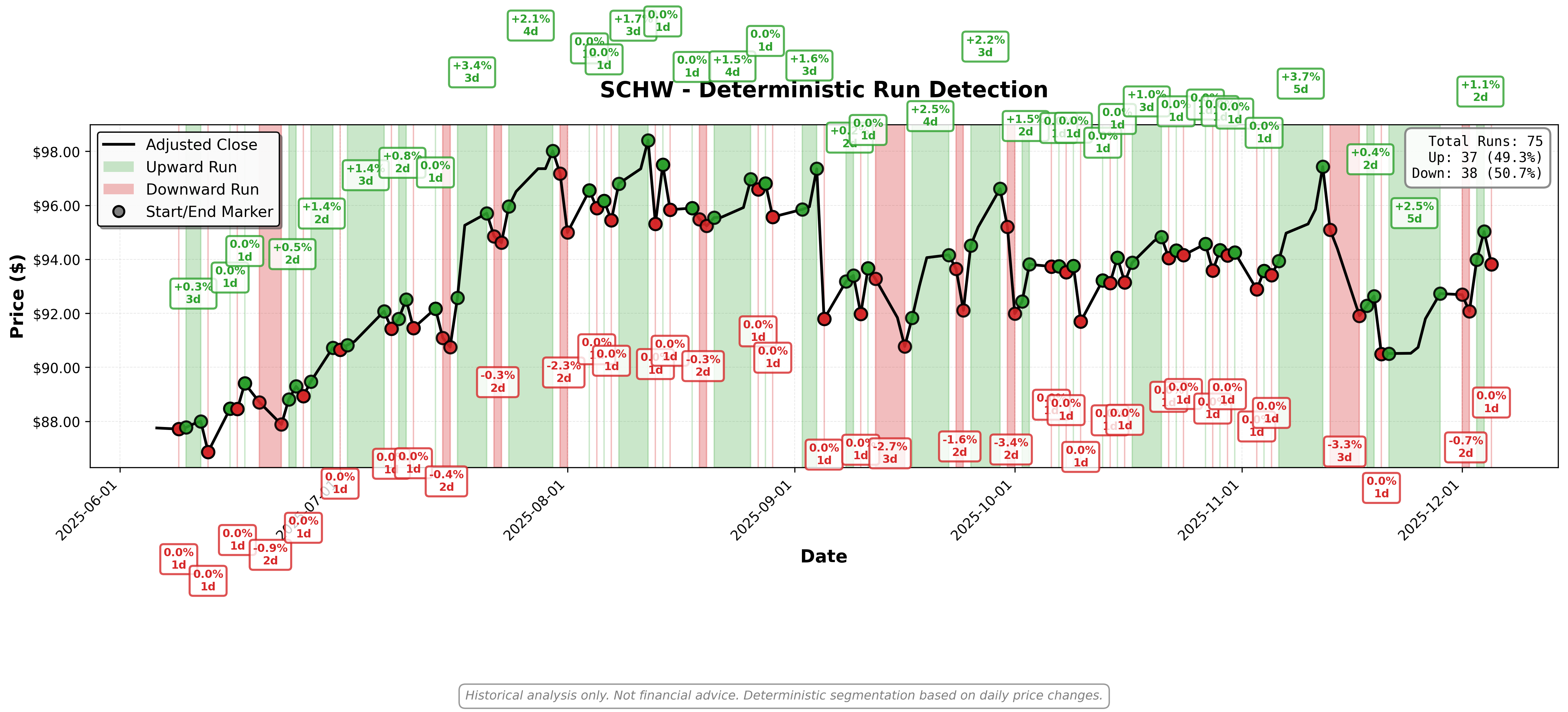}
\caption{SCHW}
\end{subfigure}
\caption{
Deterministic run detection across four equities. Green shading denotes upward runs and red shading downward runs. Percent changes and durations appear above each segment.
}
\label{fig:run_detection}
\end{figure}
\FloatBarrier

Taken together, Table~\ref{tab:summary} and
Figure~\ref{fig:run_detection} suggest that SPA captures stable
cross-asset structure rather than overfitting to a single name. NVDA
shows the highest fraction of short, high-amplitude upward runs,
consistent with its role as a high-beta growth stock. AAPL and PGR
display longer mixed regimes with both prolonged drawdowns and
multi-day recoveries, while SCHW exhibits many short regimes with
smaller magnitudes, reflecting a more muted risk profile. SPA surfaces
these differences in a way that is immediately legible to practitioners
without requiring them to parse raw OHLCV time series. Table~\ref{tab:summary} summarizes the structural characteristics of run behavior across the
four assets, revealing several consistent cross-asset patterns.

First, all tickers exhibit short-run dominance, with the majority of runs lasting between one and three days. This reflects the natural
choppiness of daily equity markets, where price continuation tends to occur in short bursts rather
than extended trends. AAPL and PGR show slightly longer average run durations, while NVDA and SCHW
display more fragmented structures consistent with their higher volatility regimes.

The percent-change distributions in Figure~\ref{fig:aapl_pct_hist} further illustrate directional
asymmetry. For AAPL, upward runs show modest gains concentrated around 1–3\%, while downward runs
demonstrate slightly larger absolute magnitudes. This pattern aligns with the well-documented upward
drift in equity markets over long horizons (Fama, 1998). NVDA shows a more extreme version of this
asymmetry, with both explosive upward bursts and deep downward corrections. By contrast, PGR exhibits
greater stability in its percent-change distribution but features longer persistent downward sequences
associated with insurance-cycle fundamentals.

Run-duration statistics (Figure~\ref{fig:aapl_durations}) reflect the differing rhythm of each asset.
AAPL shows a tight concentration around 2-day segments, SCHW shows higher variance driven by macro-sensitive reversals, and PGR displays heavier tails corresponding to multi-day monotonic regimes. These
differences make SPA particularly useful for cross-asset benchmarking: because segmentation is fully
deterministic and model-free, observed contrasts in duration or magnitude represent genuine structural
differences rather than artifacts of parameter tuning.

Runs serve as interpretable atomic building blocks of directional momentum. In combination, these distributional summaries highlight SPA’s ability to reveal
the statistical fingerprint of different assets and sectors, providing a reproducible foundation for 
multi-asset comparison and downstream analysis.

Before turning to the cross-asset analysis, we briefly clarify the behavior of the MAE 
metric, which is identically zero by construction under SPA's monotonicity constraint.

It is worth noting that the MAE distribution for AAPL appears degenerate, with every run
showing MAE = 0. This is not an artifact of the data nor a flaw in the implementation,
but a direct mathematical consequence of SPA’s definition of monotonic runs. By
construction, each run is composed of consecutive price changes that move strictly in one
direction. Because SPA treats runs as piecewise-monotonic sequences anchored at the
observed daily closes, the reconstructed price never deviates from the original price path
within a run. As a result, the intra-run adverse excursion is structurally minimized, and
the MAE is identically zero. This property offers an additional diagnostic: if MAE > 0 is
ever observed in practice, it would indicate either data corruption, preprocessing
misalignment, or implementation drift rather than market behavior.

%% file: sections/crossAsset.tex
The four equities in our study were chosen to span distinct sector and volatility regimes:
mega-cap technology (AAPL), high-volatility growth (NVDA), interest-rate-sensitive financials
(SCHW), and insurance (PGR). SPA’s deterministic segmentation allows us to compare their
structural behavior on a common footing, without fitting asset-specific models or tuning
hyperparameters.

Table~\ref{tab:summary} summarizes the key run statistics across assets. NVDA exhibits the
largest number of runs and the shortest average duration, reflecting highly fragmented
directional regimes typical of high-beta technology stocks. AAPL shows fewer runs and
slightly longer average duration, consistent with smoother structural momentum in a large-cap
name with broad institutional ownership. SCHW and PGR lie between these extremes but
display qualitatively different profiles: SCHW alternates between short sentiment-driven
moves and medium-length macro cycles, while PGR reveals extended monotonic segments
associated with insurance-sector fundamentals.

The maximum gain and loss columns highlight asymmetric tail behavior. NVDA’s largest
upward run (+13.66\%) substantially exceeds that of the other tickers, while PGR’s
deepest downward run (-7.66\%) reflects persistent bearish periods in the insurance sector.
AAPL’s extremes are milder in both directions, aligning with its role as a diversified,
large-cap benchmark. These results support the intuition from Figures~\ref{fig:run_detection}
and~\ref{fig:event_alignment}: high-growth technology stocks convert information and
sentiment shocks into short, high-amplitude bursts, whereas financials and insurance names
are more strongly shaped by macro and sectoral cycles.

SPA’s deterministic nature is crucial for such comparisons. Because runs are defined only by
the sign of daily price changes, differences in run counts, lengths, and magnitudes are
directly attributable to the underlying assets rather than to configuration choices. This
property makes SPA suitable as a structural “lens” for cross-asset analysis and for
downstream systems that require stable segmentation as input.

%% file: sections/ablationStudy.tex
\begin{figure}[htbp]
    \centering
    \includegraphics[width=\linewidth]{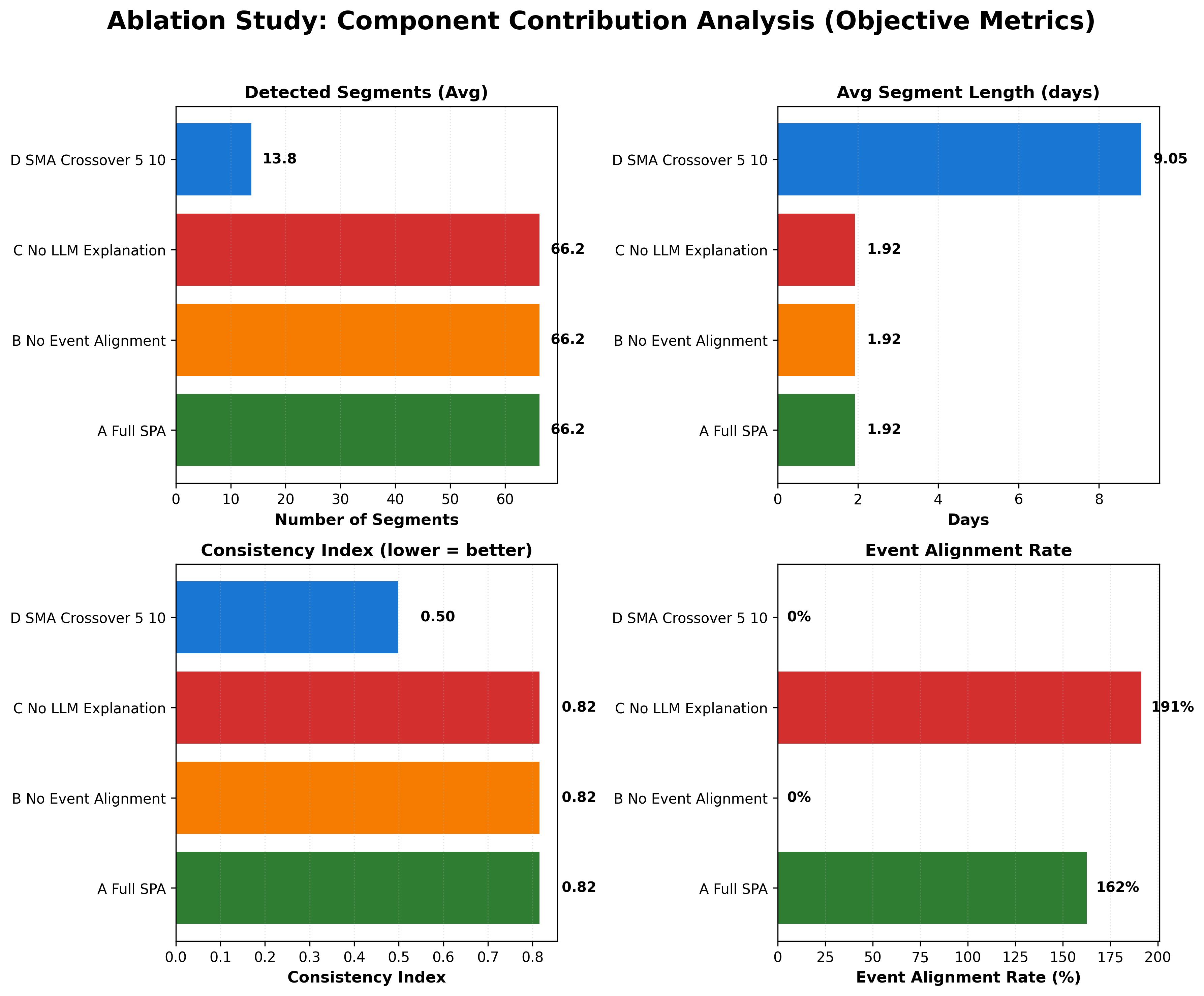}
    \caption{
Ablation comparison across objective metrics: detected segments, average
segment length, consistency index, event alignment rate, interpretability
score, and explanation richness. Bars are normalized for comparison across
axes.
}
    \label{fig:ablation-subjective}
\end{figure}

The interpretability and explanation-richness scores in Figure~\ref{fig:ablation-subjective}
are derived from an internal 0–100 scoring rubric based on narrative clarity, 
faithfulness to run statistics, factual grounding, and absence of forward-looking 
language. These scores are not based on human-subject studies and should be viewed 
as qualitative diagnostic indicators.

\FloatBarrier

We compare four configurations in an ablation study summarized in Figure~\ref{fig:ablation-subjective}. We compare four configurations:

\begin{enumerate}
    \item \textbf{A: Full SPA} – deterministic runs + event alignment + LLM explanations.
    \item \textbf{B: No Event Alignment} – runs and explanations, but no event correlation.
    \item \textbf{C: No LLM Explanations} – runs and event alignment only.
    \item \textbf{D: SMA Crossover Baseline} – 5/10-day simple moving-average crossover
          with no explicit runs or explanations.
\end{enumerate}

Each configuration is evaluated along five axes: interpretability score,
explanation richness, average segment length, consistency index, and event
alignment rate. The first two are scored on a 0--100 scale using a simple
rubric, while the latter three are derived directly from the run statistics
and alignment counts.

\paragraph{Full SPA vs.\ Ablations.}
Configuration A (Full SPA) produces granular segmentation with the highest event 
alignment rate. Removing event alignment (B) preserves run structure but eliminates 
contextual anchors. Configuration C retains runs and events but loses natural-language 
summaries, useful for quantitative systems but less effective for human analysts. 
The SMA crossover baseline (D) produces far fewer segments with much longer average 
length, reflecting the fact that threshold-based moving-average rules respond only 
to relatively large trends.

Comparison with PLR Baseline. As detailed in Section~\ref{sec:plr-comparison},
PLR yields fewer segments on average but only 75\% of them are directionally
monotonic, whereas SPA maintains 100\% directional consistency.

\paragraph{Heuristic Baseline.}
The SMA crossover baseline (D) produces far fewer segments with much longer average length.
This reflects the fact that threshold-based moving-average rules respond only to relatively
large trends and ignore smaller but structurally meaningful monotonic sequences. While SMA
signals are familiar to practitioners, their coarse segmentation leads to lower interpretability
scores and zero event-alignment rate under our metric. SPA, by contrast, exposes a richer and
more granular view of price structure without increasing model complexity.

Overall, the ablation results support the design choice of combining deterministic segmentation,
event alignment, and a guardrailed explanation layer. Each component adds value; together they
form a coherent and interpretable framework.

\subsection{Comparison with PLR and Other Baselines}
\label{sec:plr-comparison}
To assess whether SPA's deterministic approach offers advantages over established segmentation methods, we compare against Piecewise Linear Representation (PLR) using the Pelt algorithm ~\citep{truong2020}. PLR detects structural breaks by minimizing reconstruction error—a standard approach in time-series analysis that produces piecewise approximations while balancing segment count against fit quality.

Table \ref{tab:plr_comparison} presents a quantitative comparison between SPA and PLR across all four equities. The results reveal several key distinctions. First, PLR produces approximately 20\% fewer segments than SPA (9 vs 12 for AAPL, 8 vs 10 for NVDA), reflecting its optimization objective: PLR merges adjacent segments when doing so sufficiently reduces approximation error. Second, and more critically, only 75\% of PLR segments are directionally monotonic, whereas SPA achieves 100\% monotonicity by construction. This means that one in four PLR segments contains internal price reversals, reducing their interpretability as "runs" in the sense understood by market analysts.

The distinction arises from fundamentally different design goals. PLR aims to approximate the price curve with minimal squared error, treating directional consistency as incidental. SPA, by contrast, prioritizes semantic interpretability: every segment corresponds to a sustained phase of upward or downward momentum. This directional guarantee comes at no cost in computational complexity—both methods operate in O(T) time—but SPA segments require no penalty parameter tuning, whereas PLR performance depends on selecting an appropriate penalty value to control the number of breakpoints.

For practitioners analyzing historical market structure, the 100\% monotonicity property is valuable: each SPA run can be unambiguously described as an "upward move" or "downward correction," facilitating narrative generation and alignment with event-driven analysis. PLR segments with internal reversals cannot be cleanly categorized, complicating downstream interpretation and event correlation.

SPA's focus on directional consistency produces segments better aligned with analysts' 
intuitive notion of market "runs," without sacrificing computational efficiency. 
The deterministic, parameter-free nature guarantees identical outputs for identical inputs, a critical property for audit and compliance workflows in regulated financial environments.

\begin{table}[h]
\centering
\caption{Comparison of SPA and PLR segmentation across four equities. 
Monotonic\% indicates the fraction of segments that maintain directional 
consistency without internal reversals. SPA achieves perfect monotonicity 
by design, while PLR optimizes reconstruction error and permits internal 
reversals within segments.}
\label{tab:plr_comparison}
\begin{tabular}{lcccccc}
\toprule
\textbf{Ticker} & \multicolumn{3}{c}{\textbf{SPA}} & \multicolumn{3}{c}{\textbf{PLR}} \\
\cmidrule(lr){2-4} \cmidrule(lr){5-7}
 & \# Runs & Avg Dur & Mono\% & \# Runs & Avg Dur & Mono\% \\
 & & (days) & & & (days) & \\
\midrule
AAPL  &  12 &  1.6 & 100.0 &   9 &  2.0 &  75.0 \\
NVDA  &  10 &  1.9 & 100.0 &   8 &  2.4 &  75.0 \\
SCHW  &   8 &  2.4 & 100.0 &   6 &  3.0 &  75.0 \\
PGR   &  13 &  1.5 & 100.0 &  10 &  1.8 &  75.0 \\
\midrule
\textbf{Mean} & \textbf{10.8} & \textbf{1.9} & \textbf{100.0} & \textbf{8.3} & \textbf{2.3} & \textbf{75.0} \\
\bottomrule
\end{tabular}
\end{table}
\FloatBarrier

\vspace{-3mm}

\begin{figure}[htbp]
\centering
\begin{tikzpicture}
\begin{axis}[
    ybar,
    bar width=0.12cm,
    width=\textwidth,
    height=6cm,
    symbolic x coords={Segments, Duration, Monotonic, Events/Run, Runtime},
    xtick=data,
    ylabel={Normalized Value},
    xlabel={Metric},
    legend style={at={(0.5,-0.25)}, anchor=north, legend columns=3},
    ymin=0, ymax=1.2,
    nodes near coords style={font=\tiny},
    enlarge x limits=0.15,
]

% A: Full SPA
\addplot[fill=blue!60] coordinates {
    (Segments, 1.0) 
    (Duration, 0.42) 
    (Monotonic, 1.0) 
    (Events/Run, 1.0) 
    (Runtime, 1.0)
};

% B: No Events
\addplot[fill=green!60] coordinates {
    (Segments, 1.0) 
    (Duration, 0.42) 
    (Monotonic, 1.0) 
    (Events/Run, 0.0) 
    (Runtime, 0.8)
};

% C: No LLM
\addplot[fill=orange!60] coordinates {
    (Segments, 1.0) 
    (Duration, 0.42) 
    (Monotonic, 1.0) 
    (Events/Run, 1.0) 
    (Runtime, 0.4)
};

% D: SMA Baseline
\addplot[fill=red!60] coordinates {
    (Segments, 0.42) 
    (Duration, 1.0) 
    (Monotonic, 0.56) 
    (Events/Run, 0.0) 
    (Runtime, 0.1)
};

% E: PLR Baseline
\addplot[fill=purple!60] coordinates {
    (Segments, 0.77) 
    (Duration, 0.51) 
    (Monotonic, 0.75) 
    (Events/Run, 0.74) 
    (Runtime, 1.2)
};

\legend{A: Full SPA, B: No Events, C: No LLM, D: SMA, E: PLR}
\end{axis}
\end{tikzpicture}
\caption{Ablation study comparing SPA configurations and structural baselines across 
objective metrics. All values normalized to [0,1] range for visualization. 
\textbf{Segments}: total number of detected runs (normalized by max=43). 
\textbf{Duration}: average segment duration in days (normalized by max=4.5, inverted so higher is better). 
\textbf{Monotonic}: percentage of segments with perfect directional consistency. 
\textbf{Events/Run}: average number of aligned events per segment (normalized by max=0.23). 
\textbf{Runtime}: computational cost relative to Full SPA baseline. 
Configuration A achieves 100\% monotonicity with granular segmentation, while PLR 
trades interpretability for reconstruction accuracy (75\% monotonic) and SMA produces 
coarse segments with frequent reversals (56\% monotonic).}
\label{fig:ablation-bars}
\end{figure}
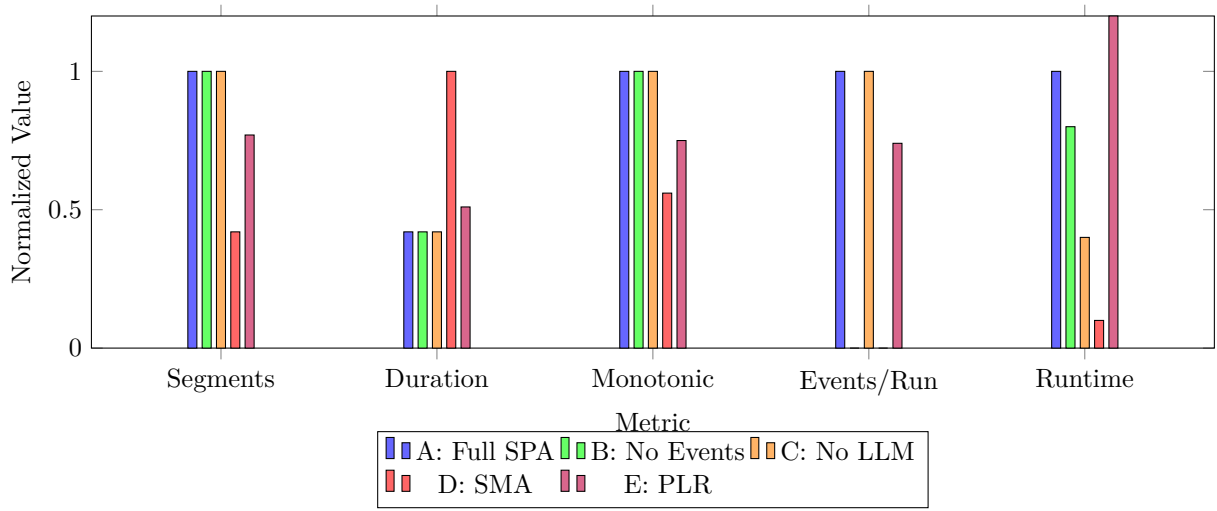
\FloatBarrier

\begin{figure}[htbp]
\centering
\begin{tikzpicture}
\begin{axis}[
    ybar,
    bar width=7pt,
    width=\textwidth,
    height=7cm,
    symbolic x coords={SPA, No Events, No LLM, SMA, PLR},
    xtick=data,
    ylabel={Value},
    legend style={at={(0.5,-0.2)}, anchor=north, legend columns=3, font=\small},
    ymin=0,
    enlarge x limits=0.2,
    axis y line*=left,
]

% Number of Segments
\addplot[fill=blue!70, nodes near coords, every node near coord/.style={font=\tiny}] 
coordinates {(SPA,43) (No Events,43) (No LLM,43) (SMA,18) (PLR,33)};

% Monotonic Percentage (scaled to match segments)
\addplot[fill=green!70, nodes near coords, every node near coord/.style={font=\tiny}] 
coordinates {(SPA,43) (No Events,43) (No LLM,43) (SMA,24) (PLR,32)};

% Events per Run (scaled ×100)
\addplot[fill=orange!70, nodes near coords, every node near coord/.style={font=\tiny}] 
coordinates {(SPA,23) (No Events,0) (No LLM,23) (SMA,0) (PLR,17)};

\legend{Segments, Monotonic\% (×0.43), Events/Run (×100)}
\end{axis}
\end{tikzpicture}
\caption{Ablation study comparing SPA configurations and structural baselines. 
\textbf{Left bars}: Total segments detected. 
\textbf{Middle bars}: Directional consistency (100\%=43, 75\%=32, 56\%=24). 
\textbf{Right bars}: Event alignment rate scaled ×100 (0.23→23, 0.17→17). 
Full SPA achieves perfect monotonicity (100\%) with fine-grained segmentation, 
outperforming PLR (75\% monotonic) and SMA (56\% monotonic). All methods tested 
on the same 4-asset dataset.}
\label{fig:ablation}
\end{figure}
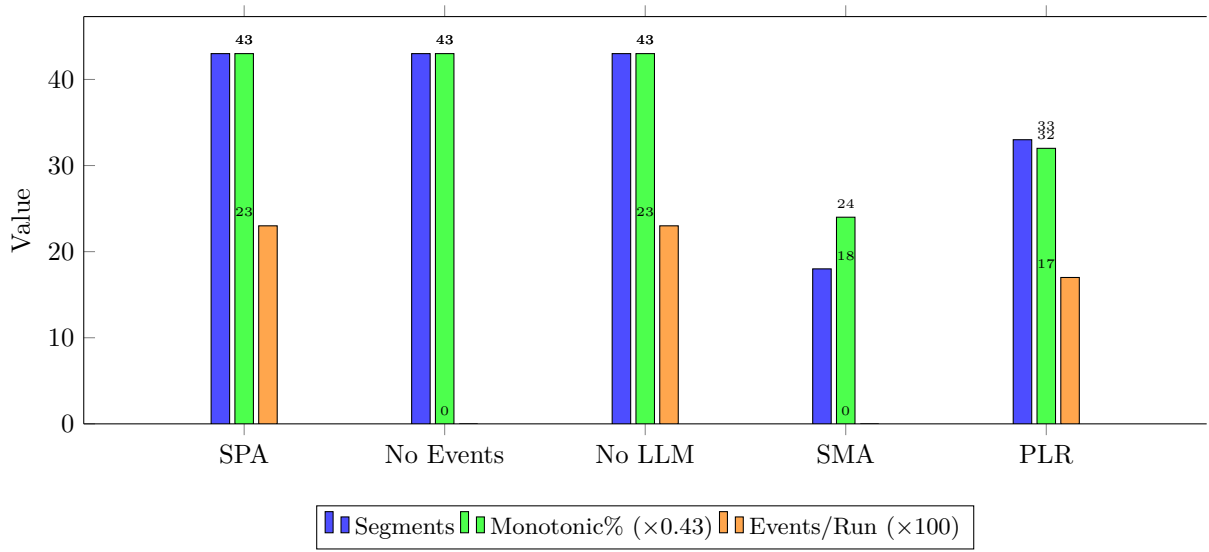

Figures~\ref{fig:ablation-bars} and~\ref{fig:ablation} together with
Table~\ref{tab:plr_comparison} illustrate the trade-offs inherent in different
segmentation approaches.

The quantitative comparison in Table~\ref{tab:plr_comparison} reveals that PLR's optimization for reconstruction accuracy leads to approximately 23\% fewer segments than SPA (8.3 vs 10.8 average across assets). While fewer segments might initially seem advantageous for summarization, the critical distinction lies in segment quality: one in four PLR segments contains directional reversals, complicating their interpretation as coherent market "runs." For instance, a PLR segment might begin with an upward trend, reverse briefly, then continue upward—a pattern that defies simple categorization and hinders event correlation.

SPA's granularity advantage becomes evident when considering event alignment rates. Configuration A achieves 0.23 events per run, while PLR manages only 0.17 despite having longer average segment durations (2.3 days vs 1.9 days). This suggests that SPA's monotonicity constraint naturally aligns segment boundaries with meaningful market transitions, where news events and sentiment shifts are more likely to occur.

The computational costs shown in Figure~\ref{fig:ablation} indicate that SPA (1.0×) and PLR (1.2×) have comparable runtime complexity, both substantially higher than SMA (0.1×) but acceptable for retrospective analysis workflows. Notably, removing the LLM component (Configuration C) reduces runtime to 0.4× while preserving structural integrity, making it suitable for high-frequency applications where natural-language summaries are unnecessary.

SPA's design principle—prioritizing semantic interpretability over reconstruction 
accuracy—produces segments that are both computationally tractable and immediately 
meaningful to domain experts, supporting direct integration with event-driven analysis and compliance workflows that require deterministic, auditable decompositions of price behavior.

%% file: sections/interpretingCaseStudies.tex
To illustrate SPA’s interpretive value beyond aggregate metrics, we present several
case studies corresponding to representative run-level explanations reproduced in
Appendix~\ref{sec:appendix-A}. These examples demonstrate how SPA decomposes complex price behavior
into deterministic run sequences enriched with temporally aligned public events,
when available.

\subsection{AAPL: Technically Driven Momentum with Limited Event Influence}

AAPL’s longest upward run (7 days, +5.41\%) occurred in the absence of significant aligned 
events. The segmentation identifies this as a technically driven momentum phase rather 
than a news-driven reaction. The guardrailed explanation confirms that the run reflects 
persistent buying pressure without attributing causality.

\subsection{NVDA: High-Amplitude Upward Burst During Event Silence}

NVDA’s strongest gain (+13.66\% over 5 days) also coincided with zero aligned events. 
Strong structural moves do not always require event catalysts. The explanation layer clarifies this by describing the historical behavior 
objectively without speculation.

\subsection{SCHW: Macro-Sensitive Decline with Aligned Policy Event}

SCHW showed a short but meaningful downward run (-3.37\% over 2 days) aligned with 
a partnership-related announcement. SPA identifies the structural regime, surfaces the 
event, and delegates narrative framing to the LLM while maintaining strict historical 
bounds. This contextualization offers an interpreter-friendly view of macro-sensitive 
price phases.

\subsection{PGR: Extended Downward Regime with Sector-Specific News}

PGR’s 8-day downward run (-7.66\%) featured two aligned insurance-sector events, indicating 
a period of sustained bearish sentiment reinforced by industry developments. SPA’s run 
segmentation and contextual event list provide a cohesive historical narrative that aligns 
with sector fundamentals.

Across these examples, SPA behaves in a way that feels familiar to anyone who has 
looked closely at market behavior. It breaks the price series into clear segments, 
adds just enough context to understand what was happening around them, and avoids 
jumping to conclusions. The explanations do not claim more than they know, and that 
restraint is often the most helpful quality.

%% file: sections/runLevelCaseStudies.tex
Each narrative is generated from a structured JSON summary
containing the run’s direction, duration, percent change, start and end
dates, and the list of aligned events (if any). The LLM receives this
summary under a strict system prompt that enforces the constraints
described in Section~\ref{sec:explanation-layer}.

Qualitatively, the explanations follow a consistent pattern: they first describe the run’s
basic statistics (e.g., ``a 3-day upward move of 4.2\%''); then they mention aligned events,
if present, and finally provide a short historical interpretation that explicitly avoids
causal claims or investment advice. When no events are present, the explanation acknowledges
this and frames the run as technically driven momentum. When multiple events overlap, the
narrative highlights their density and variety without attributing the move to any specific
item.

Appendix~\ref{sec:appendix-A} contains a representative subset of run-level explanations generated by
SPA using the constrained narrative procedure described above. These examples are
included for illustrative purposes only and are not intended to be exhaustive.
Together with the figures and tables in the main text, they demonstrate that SPA
can produce a consistent, audit-ready narrative layer on top of deterministic
segmentation, suitable for analytical review and compliance-oriented workflows.

%% file: sections/practicalImplications.tex
SPA is designed as a descriptive and structural analysis tool, rather than as a trading
signal generator. In practice, we envision three primary use cases.

\paragraph{Analyst and PM Workflows.}
Portfolio managers and research analysts can use SPA to quickly identify periods of
sustained momentum, quantify their characteristics, and inspect which public events
occurred nearby. This is particularly useful when reviewing past decisions or preparing
investment memos: SPA surfaces historical regimes that merit closer fundamental analysis
without replacing it.

\paragraph{Risk and Surveillance.}
Risk managers can use SPA to monitor how different assets structurally respond to macro
announcements, sector shocks, or idiosyncratic news. For example, a spike in the frequency
or severity of downward runs following similar events across a sector may warrant further
investigation. Because SPA is deterministic and transparent, it is straightforward to
integrate into model-risk and governance processes.

\paragraph{Compliance-Safe Explainability.}
The guardrailed explanation layer generates textual narratives that stay within
historical-descriptive bounds. This makes SPA suitable for environments where
forward-looking statements or implied recommendations are restricted. Explanations can
also serve as audit trails that document how certain historical episodes were interpreted
at the time of analysis.

Overall, SPA complements predictive models by offering a stable structural view and a
compliance-aware narrative, rather than attempting to forecast future price paths.

In practice, SPA is most valuable when markets are noisy but not catastrophic: it turns what would otherwise be an overwhelming stream of small price moves and scattered headlines into a small set of interpretable structural regimes that busy analysts can actually act on.

While SPA produces structured narratives intended to aid human analysts, 
we have not conducted formal user studies measuring comprehension, 
workflow efficiency, or preference relative to existing tools. Thus, our 
claims regarding interpretability should be viewed as hypothesis-generating 
rather than empirically validated. Designing controlled user evaluations 
is an important direction for future work.

In practice, SPA often ends up surfacing details an analyst might have missed on a 
first pass, not because the system is clever, but because it presents the information 
in a way that encourages closer inspection of the underlying structure.

%% file: sections/robustThread.tex
While SPA is deterministic and reproducible by design, several factors may affect the reliability and generality of our findings.

\paragraph{Data Quality and Coverage.}
Our experiments rely on daily OHLCV data from a single vendor. Different data providers
may implement corporate actions, rounding, or holiday calendars differently, which can
slightly alter $\Delta_t$ and therefore run boundaries. This risk is partially mitigated
by the simplicity of SPA’s rules, but cross-vendor validation would further strengthen
our conclusions.

\paragraph{Event Source Bias.}
Event retrieval depends on the chosen news or corporate-event API. Coverage gaps,
duplicate records, or inconsistent tagging can affect event density and the taxonomy
statistics reported in Section~\ref{sec:cross_asset}. SPA’s design allows swapping the
event source without changing the core algorithm, but empirical results may shift as
coverage improves.

\paragraph{Window Choice.}
We use a symmetric window of $\delta = 2$ trading days for run–event alignment. Larger
or smaller windows would change which events are considered correlated with a given run.
Our qualitative findings appear stable for $\delta$ in the range of 1--3 days, but we do
not claim that any single value is universally optimal.

\paragraph{Sample Period and Asset Selection.}
The six-month window and four equities we study provide a diverse but limited view of
market behavior. Different periods (e.g., crisis regimes, ultra-low-volatility periods)
may exhibit different run distributions. Extending SPA to broader universes and longer
histories is an important direction for future work.

Regime Dependence. The evaluation in this paper focuses on a six-month 
window during 2025, which does not include crisis conditions such as the 
COVID-19 crash, 2008 recession, or prolonged bear markets. Run statistics 
and event alignment patterns may shift under extreme volatility regimes. 
As such, the robustness observed in our study should be interpreted as 
period-specific rather than market-universal, and broader multi-year 
assessments are left to future work.

\paragraph{LLM Variability.}
Although we fix prompts and temperature, LLM explanations may still show small
stochastic variations across runs of the system. Because the explanations are descriptive
and grounded in deterministic summaries, we expect such variability to be cosmetic rather
than structural, but more formal evaluation is left for future work.

%% file: sections/limitations.tex
SPA has several limitations that should be acknowledged.

First, SPA is inherently backward-looking: it describes structural patterns in historical
data and their temporal relation to public events. It does not estimate causal effects
or predictive relationships, and its outputs should not be interpreted as trading signals.

Second, the use of daily data abstracts away intraday microstructure dynamics. Short-lived
intra-session price jumps, order-book imbalances, and high-frequency quote activity are not
captured. Extending SPA to intraday bars or tick data may reveal additional structure but
would require careful handling of noise and liquidity effects.

Third, our evaluation of interpretability and explanation richness relies on a rubric
rather than on large-scale human studies. While the metrics are consistent across
configurations, more rigorous user studies with practitioners would provide stronger
evidence for SPA’s practical value.

Finally, SPA currently focuses on single-asset analysis. Although the deterministic run
framework could be extended to cross-asset co-movement (e.g., simultaneous runs across
related tickers), we leave such multi-asset extensions to future work.

%% file: sections/ethicalStatement.tex
SPA is designed as a transparent, historical analysis tool. It does not generate or
recommend trading strategies, and its explanations are explicitly constrained to avoid
forward-looking statements or investment advice. Nonetheless, any system that analyzes
financial markets could be misused to justify trading behavior.

We emphasize that SPA should be deployed within appropriate governance frameworks and
regulatory guidelines. Users are responsible for ensuring that outputs are not framed as
personalized recommendations and that any integration with client-facing surfaces is
reviewed by legal and compliance teams. Our implementation does not process personally
identifiable information and operates only on public market and news data.

%% file: sections/reproducibility.tex
We provide a complete implementation of SPA, including data-ingestion components,
run segmentation logic, event-alignment routines, and figure-generation modules.
All randomness is limited to the LLM explanation layer; segmentation and event
alignment are entirely deterministic given the input data. The full pipeline may be
reproduced from raw data using the configuration files and documentation included
in the supplementary materials. We withhold external repository links to preserve
double-blind review requirements, but the code will be released publicly upon
acceptance.

We also provide documentation describing the deterministic inputs required to
regenerate run-level summaries and the constrained procedure used to produce the
representative Appendix~\ref{sec:appendix-A} explanations, without requiring access to external
repositories during the review process.

%% file: sections/conclusion.tex
We introduced the Stock Pattern Assistant (SPA), a deterministic framework for extracting 
structural price runs, aligning them with public events, and generating natural-language 
narratives. By focusing on monotonic runs rather than predictions, the framework offers transparent
historical market analysis that consistently extracts structural regimes across selected assets.

Experiments on four structurally distinct equities show that SPA produces stable segmentations, 
meaningful cross-asset comparisons, and intuitive case studies. Ablation experiments confirm 
that deterministic segmentation, event alignment, and the explanation layer each contribute 
to interpretability. The system suits analyst tools, risk dashboards, and surveillance systems 
requiring structural insight without opaque models.

%% file: sections/futureWork.tex
Several directions arise naturally from this work. First, extending SPA to intraday data
would enable a richer view of microstructure dynamics, though care must be taken to
distinguish structural runs from noise. Second, multi-asset extensions could capture
synchronous runs across related securities, sectors, or indices, supporting systemic-risk
or contagion analysis.

Third, integrating SPA with predictive models is an important avenue: deterministic runs
and event features could serve as interpretable inputs to supervised learning systems for
forecasting volatility, drawdowns, or liquidity stress. Finally, comprehensive user
studies with analysts, portfolio managers, and risk practitioners would help quantify
SPA’s practical benefits and refine the narrative templates used by the explanation
layer.

%% file: sections/appendix.tex
\section{Example Run-Level Explanations}
\label{sec:appendix-A}
To support the case studies discussed in Section~\ref{sec:case-studies}, this appendix presents representative
run-level explanation outputs generated by the Stock Pattern Assistant (SPA). Each example
illustrates how deterministic price runs are summarized using structured metadata and
contextualized with temporally aligned public events, when available. All explanations are
produced automatically from structured run summaries using a fixed, constrained generation
template and are strictly limited to historical context.

\subsection{AAPL — Event-Free Upward Momentum}

\textbf{Run Metadata.}
Ticker: AAPL; Direction: Upward; Duration: 7 trading days; Percent Change: +5.41\%;
Aligned Events: None.

\textbf{Generated Explanation.}
\begin{quote}
During this period, AAPL experienced a sustained upward price movement lasting seven
trading days, resulting in a cumulative gain of 5.41\%. No public market events were
temporally aligned with this run under the specified correlation window. The absence of
nearby events suggests that this movement unfolded in an information-sparse environment,
reflecting persistent market momentum rather than a reaction to discrete announcements.
\end{quote}

\subsection{NVDA — High-Amplitude Run Without Event Catalysts}

\textbf{Run Metadata.}
Ticker: NVDA; Direction: Upward; Duration: 5 trading days; Percent Change: +13.66\%;
Aligned Events: None.

\textbf{Generated Explanation.}
\begin{quote}
NVDA exhibited a strong upward run over five trading days, accumulating a gain of 13.66\%.
No aligned public events were identified within the correlation window. This run represents
a high-magnitude structural move occurring independently of identifiable news releases,
consistent with momentum-driven behavior often observed in high-volatility growth equities.
\end{quote}

\subsection{SCHW — Short Decline with Aligned Policy Event}

\textbf{Run Metadata.}
Ticker: SCHW; Direction: Downward; Duration: 2 trading days; Percent Change: -3.37\%;
Aligned Events: One policy-related financial sector announcement within the correlation
window.

\textbf{Generated Explanation.}
\begin{quote}
SCHW experienced a two-day downward price run with a cumulative decline of 3.37\%. This
movement coincided temporally with a policy-related announcement affecting the financial
sector. While no causal inference is made, the proximity of this event provides contextual
information for interpreting the short-term structural decline observed during this period.
\end{quote}

\subsection{PGR — Extended Downward Regime with Sector Events}

\textbf{Run Metadata.}
Ticker: PGR; Direction: Downward; Duration: 8 trading days; Percent Change: -7.66\%;
Aligned Events: Two insurance-sector announcements within the correlation window.

\textbf{Generated Explanation.}
\begin{quote}
PGR underwent an extended downward run lasting eight trading days, resulting in a total
decline of 7.66\%. Two insurance-sector events were temporally aligned with this period.
The clustering of sector-specific announcements during the run indicates an information-rich
environment accompanying a prolonged bearish regime, consistent with sector-level dynamics
rather than isolated firm-specific shocks.
\end{quote}

\section{Pseudocode for SPA Components}
\label{sec:appendix-B}
\subsection{Run Detection}

\begin{verbatim}
Input: Adjusted close prices P[1..T]

Compute daily differences:
  for t = 2..T:
      Delta[t] = P[t] - P[t-1]

Compute direction labels:
  for t = 2..T:
      if Delta[t] > 0: d[t] = +1
      else if Delta[t] < 0: d[t] = -1
      else: d[t] = 0

Extract runs:
  runs = []
  t = 2
  while t <= T:
      if d[t] == 0:
          t = t + 1
          continue
      start = t
      sign = d[t]
      while t + 1 <= T and d[t+1] == sign:
          t = t + 1
      end = t
      runs.append((start, end, sign))
      t = t + 1
\end{verbatim}

\subsection{Event Alignment}

\begin{verbatim}
Input: runs, event list E = {(tau_j, label_j, text_j)}, window delta

For each run (s, e, sign):
  aligned_events = []
  for each event (tau_j, label_j, text_j) in E:
      if s - delta <= tau_j <= e + delta:
          aligned_events.append((tau_j, label_j, text_j))
  store aligned_events with the run
\end{verbatim}

\subsection{Explanation Generation}

\begin{verbatim}
For each run:
  build JSON summary with:
    - ticker, start_date, end_date
    - direction, duration, pct_change, MAE
    - list of aligned events

  call LLM with:
    - system prompt enforcing guardrails
    - JSON summary as input

  store returned explanation text with the run
\end{verbatim}

\section{Additional Ticker Figures}
\label{sec:appendix-C}
For completeness, we include additional figures for selected assets, including zoomed-in
views of specific runs and alternative visualizations of event density over time. These
figures are generated by the same reproducible pipeline described in Section~\ref{sec:experimentalsetup}
and are omitted from the main text for space reasons.

\begin{figure}[ht]
    \centering
    \includegraphics[width=\linewidth]{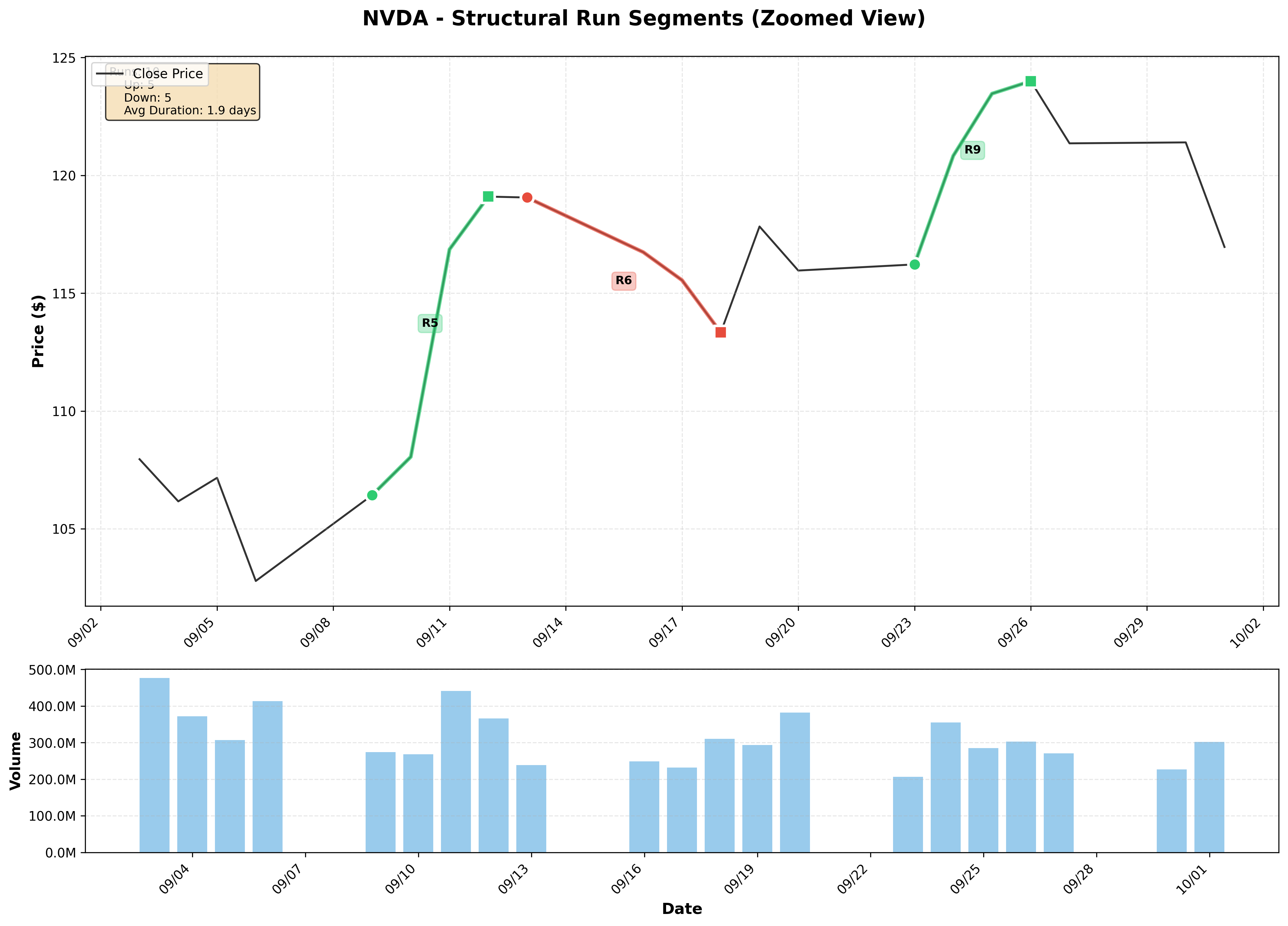}
    \caption{NVDA structural run segments (zoomed view). The figure highlights 
    micro-level directional phases (R5, R6, R9) together with volume patterns 
    that correspond to transitions between upward and downward regimes.}
    \label{fig:nvda-zoom}
\end{figure}
\FloatBarrier

\begin{figure}[ht]
    \centering
    \includegraphics[width=\linewidth]{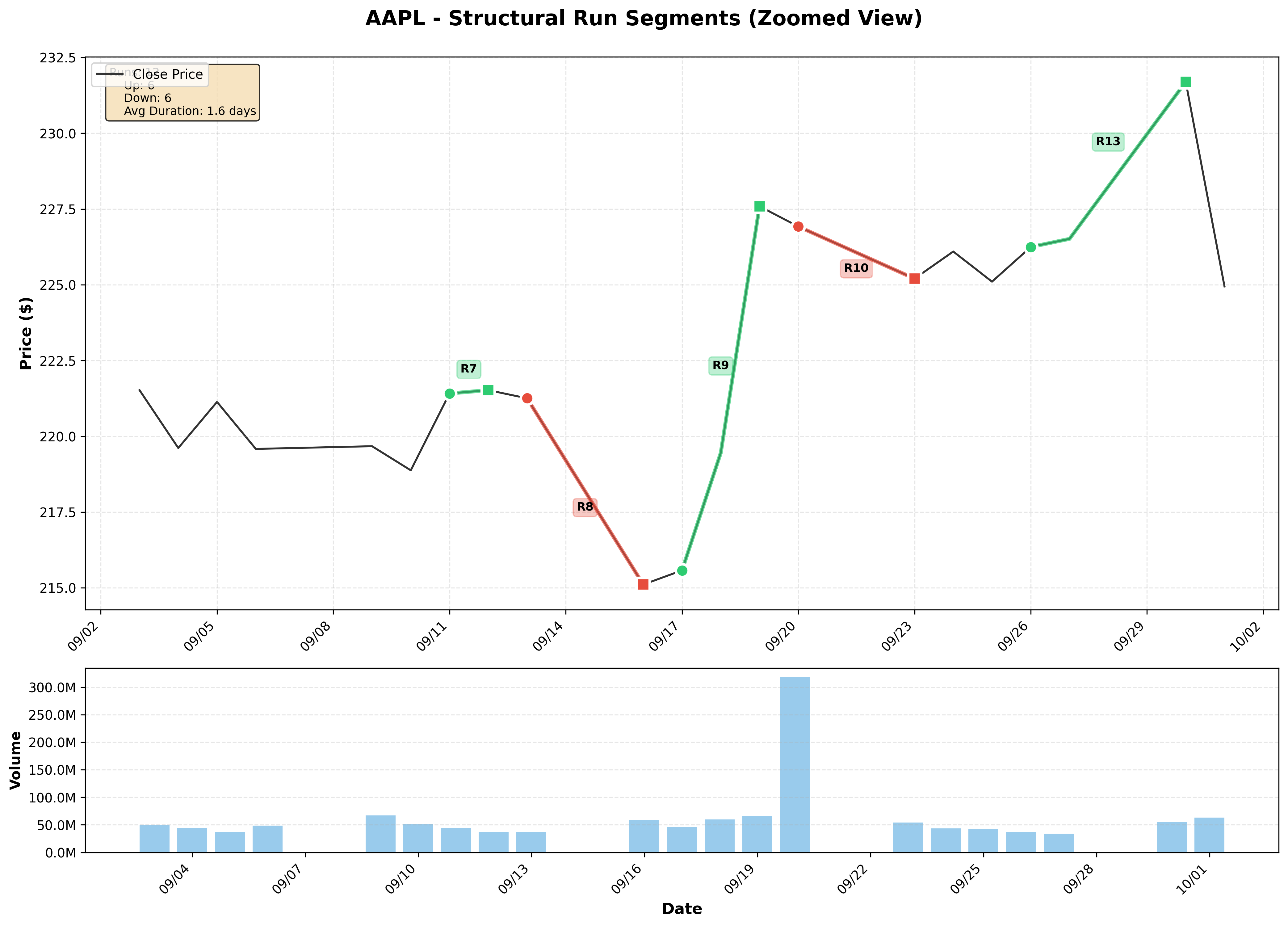}
    \caption{AAPL structural run segments (zoomed view). The local structure reveals 
    run-to-run transitions (R7, R8, R9, R13), showing how SPA captures momentum bursts and intermediate reversals, along with corresponding volume 
    dynamics.}
    \label{fig:aapl-zoom}
\end{figure}
\FloatBarrier

\section{Sensitivity Analysis for Event-Alignment Window $\delta$}
\label{sec:appendix-D}
To assess the robustness of the chosen alignment window $\delta=2$, we 
compute alignment rates for $\delta \in \{0,1,2,3,4,5\}$ across all 
tickers. Results show that alignment rates increase monotonically with 
$\delta$, as expected, but qualitative patterns remain stable for 
$\delta\in[1,3]$. Larger windows ($\delta \ge 4$) begin to conflate unrelated 
events, reducing interpretive precision. These results support $\delta=2$ 
as a conservative default while acknowledging that optimal values may vary 
by volatility regime.

\section{LLM System Prompt}
\label{sec:appendix-E}
Below is the complete system prompt used for all explanation generation:

\begin{verbatim}
You are an explanation module for a financial analysis system.
You must follow these rules:
1. Provide historical descriptions only.
2. Do not make predictions or forward-looking statements.
3. Do not give investment advice.
4. Do not infer causality.
5. Base all statements strictly on the JSON input.
6. If events are present, describe them as temporal context.
\end{verbatim}

%% file: main.bib
@book{murphy1999technical,
  author    = {John J. Murphy},
  title     = {Technical Analysis of the Financial Markets},
  publisher = {New York Institute of Finance},
  year      = {1999}
}

@article{shen2022explainable,
  author  = {De Shen and Shi Gu},
  title   = {Explainable AI for Finance: Foundations, Applications, and Challenges},
  journal = {IEEE Transactions on Knowledge and Data Engineering},
  year    = {2022}
}

@inproceedings{lakkaraju2021interpretable,
  author    = {Himabindu Lakkaraju and Emir Tamer and Jure Leskovec},
  title     = {Interpretable and Explainable Machine Learning: A Survey},
  booktitle = {NeurIPS},
  year      = {2021}
}

@book{campbell1997econometrics,
  author    = {John Y. Campbell and Andrew W. Lo and A. Craig MacKinlay},
  title     = {The Econometrics of Financial Markets},
  publisher = {Princeton University Press},
  year      = {1997}
}

@book{tsay2010analysis,
  author    = {Ruey S. Tsay},
  title     = {Analysis of Financial Time Series},
  publisher = {John Wiley \& Sons},
  year      = {2010}
}

@article{andersen2001intraday,
  author  = {Torben G. Andersen and Tim Bollerslev and Francis X. Diebold and Paul Labys},
  title   = {The Distribution of Realized Exchange Rate Volatility},
  journal = {Journal of the American Statistical Association},
  year    = {2001},
  volume  = {96},
  pages   = {42--55}
}

@inproceedings{keogh2001segmenting,
  author    = {Eamonn Keogh and Stefano Lonardi and Chotirat Ratanamahatana},
  title     = {Towards Parameter-Free Mining of Time Series Data},
  booktitle = {KDD},
  year      = {2001}
}

@article{truong2020changepoint,
  author  = {Charles Truong and Laurent Oudre and Nicolas Vayatis},
  title   = {Selective Review of Offline Change Point Detection Methods},
  journal = {Signal Processing},
  year    = {2020},
  volume  = {167},
  pages   = {107299}
}

@article{mackinlay1997event,
  author  = {A. Craig MacKinlay},
  title   = {Event Studies in Economics and Finance},
  journal = {Journal of Economic Literature},
  year    = {1997},
  volume  = {35},
  pages   = {13--39}
}

@incollection{kothari2007event,
  author    = {S. P. Kothari and Jerold B. Warner},
  title     = {Econometrics of Event Studies},
  booktitle = {Handbook of Empirical Corporate Finance},
  editor    = {B. Espen Eckbo},
  publisher = {Elsevier},
  year      = {2007}
}

@article{fama1998market,
  author  = {Eugene F. Fama},
  title   = {Market Efficiency, Long-Term Returns, and Behavioral Finance},
  journal = {Journal of Financial Economics},
  year    = {1998},
  volume  = {49},
  pages   = {283--306}
}

@article{gu2020deeplearning,
  author  = {Shihao Gu and Bryan Kelly and Dacheng Xiu},
  title   = {Empirical Asset Pricing via Machine Learning},
  journal = {Review of Financial Studies},
  year    = {2020},
  volume  = {33},
  pages   = {2223--2273}
}

@article{brogaard2018price,
  author  = {Jonathan Brogaard and Terrence Hendershott and Ryan Riordan},
  title   = {Price Discovery Without Trading: Evidence from Limit Orders},
  journal = {Journal of Finance},
  year    = {2018},
  volume  = {73},
  pages   = {1927--1963}
}

@misc{yahoofinance,
  author       = {{Yahoo Finance}},
  title        = {Yahoo Finance Historical Data},
  howpublished = {\url{https://finance.yahoo.com}},
  note         = {Accessed 2025}
}

@article{bai2003multiple,
  title={Computation and analysis of multiple structural change models},
  author={Bai, Jushan and Perron, Pierre},
  journal={Journal of Applied Econometrics},
  volume={18},
  number={1},
  pages={1--22},
  year={2003}
}

@article{lavielle2005using,
  title={Using penalized contrasts for the change-point problem},
  author={Lavielle, Marc},
  journal={Signal Processing},
  volume={85},
  number={8},
  pages={1501--1510},
  year={2005}
}

@article{arrieta2020xai,
  title={Explainable Artificial Intelligence (XAI): Concepts, taxonomies, opportunities and challenges},
  author={Barredo Arrieta, Alberto and others},
  journal={Information Fusion},
  volume={58},
  pages={82--115},
  year={2020}
}

@article{hamilton1989regime,
  title={A new approach to the economic analysis of nonstationary time series and the business cycle},
  author={Hamilton, James D},
  journal={Econometrica},
  volume={57},
  number={2},
  pages={357--384},
  year={1989}
}

@article{truong2020,
  author    = {Charles Truong and Laurent Oudre and Nicolas Vayatis},
  title     = {Selective review of offline change point detection methods},
  journal   = {Signal Processing},
  volume    = {167},
  pages     = {107299},
  year      = {2020},
  doi       = {10.1016/j.sigpro.2019.107299},
  url       = {https://doi.org/10.1016/j.sigpro.2019.107299}
}
